\documentclass{article}

\usepackage{microtype}
\usepackage{amsfonts}
\usepackage{amsmath}
\usepackage{graphicx}
\usepackage{multicol, multirow}   
\usepackage{subcaption}
\usepackage{mwe}
\usepackage{booktabs} 
\newbox{\bigpicturebox}

\usepackage{hyperref}
\usepackage{comment}
\usepackage[dvipsnames]{xcolor}

\usepackage[accepted]{icml2021}

\icmltitlerunning{Loss Surface Simplexes}

\usepackage{selectp}

\begin{document}

\twocolumn[
\icmltitle{Loss Surface Simplexes for Mode Connecting Volumes and Fast Ensembling}

\begin{icmlauthorlist}
\icmlauthor{Gregory W. Benton}{nyu}
\icmlauthor{Wesley J. Maddox}{nyu}
\icmlauthor{Sanae Lotfi}{nyu}
\icmlauthor{Andrew Gordon Wilson}{nyu}
\end{icmlauthorlist}

\icmlaffiliation{nyu}{New York University}

\icmlcorrespondingauthor{Gregory W. Benton}{gwb260@nyu.edu}

\icmlkeywords{Machine Learning, ICML}

\vskip 0.3in
]

\printAffiliationsAndNotice{} 

\begin{abstract}
With a better understanding of the loss surfaces for multilayer networks, we can build more robust and accurate training procedures. 
Recently it was discovered that independently trained SGD solutions can be connected along one-dimensional paths of near-constant training loss. 
In this paper, we show that there are in fact mode-connecting simplicial complexes that form multi-dimensional manifolds of low loss, connecting many independently trained models. 
Inspired by this discovery, we show how to efficiently build simplicial complexes for fast ensembling, outperforming independently trained deep ensembles in accuracy, calibration, and robustness to dataset shift. 
Notably, our approach only requires a few training epochs to discover a low-loss simplex, starting from a pre-trained solution.
Code is available at \url{https://github.com/g-benton/loss-surface-simplexes}.
\end{abstract}

\section{Introduction}
\label{sec: intro}

Despite significant progress in the last few years, little is known about neural network loss landscapes.
Recent works have shown that the modes found through SGD training of randomly initialized models are connected along narrow pathways connecting pairs of modes, or through tunnels connecting multiple modes at once \citep{garipov2018loss, draxler2018essentially, fort2019large}. In this paper we show that there are in fact large multi-dimensional simplicial complexes of low loss in the parameter space of neural networks that contain arbitrarily many independently trained modes.

\begin{figure}
\centering
\includegraphics[width=\linewidth,trim={0 12cm 0 5cm},clip]{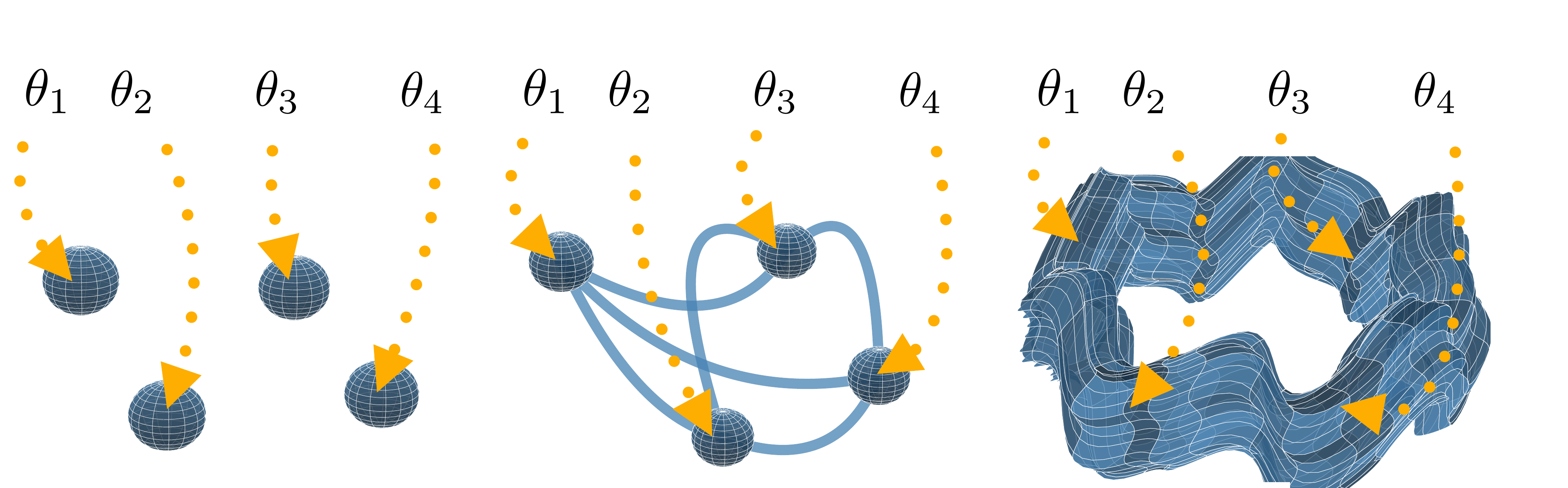}
\caption{A progressive understanding of the loss surfaces of neural networks. 
\textbf{Left:} The traditional view of loss in parameter space, in which regions of low loss are disconnected \citep{goodfellow2014qualitatively,choromanska2015loss}. 
\textbf{Center:} The revised view of loss surfaces provided by work on mode connectivity; multiple SGD training solutions are connected by narrow tunnels of low loss \citep{garipov2018loss,draxler2018essentially,fort2019large}.
\textbf{Right:} The viewpoint introduced in this work; SGD training converges to different points on a connected \emph{volume} of low loss. Paths between different training solutions exist within a large multi-dimensional manifold of low loss. We provide a two dimensional representation of these loss surfaces in Figure \ref{fig: example-2D}.}
\label{fig:loss_surface_theoretical}
\end{figure}

The ability to find these large volumes of low loss that can connect any number of independent training solutions represents a natural progression in how we understand the loss landscapes of neural networks, as shown in Figure \ref{fig:loss_surface_theoretical}. 
In the left of Figure \ref{fig:loss_surface_theoretical}, we see the classical view of loss surface structure in neural networks, where there are many isolated low loss modes that can be found through training randomly initialized networks. In the center we have a more contemporary view, showing that there are paths that connect these modes. On the right we present a new view --- that all modes found through standard training  converge to points within a single connected \emph{multi-dimensional} volume of low loss. 

We introduce Simplicial Pointwise Random Optimization (SPRO) as a method of finding simplexes and simplicial complexes that bound volumes of low loss in parameter space. With SPRO we are able to find mode connecting spaces that simultaneously connect many independently trained models through a a single well-defined multi-dimensional manifold. Furthermore, SPRO is able to explicitly define a space of low loss solutions through determining the bounding vertices of the simplicial complex, meaning that computing the dimensionality and volume of the space become straightforward, as does sampling models within the complex.

This enhanced understanding of loss surface structure enables practical methodological advances.
Through the ability to rapidly sample models from within the simplex we can form Ensembled SPRO (ESPRO) models. ESPRO works by generating a simplicial complex around independently trained models and ensembling from within the simplexes, outperforming the gold standard deep ensemble combination of independently trained models \citep{lakshminarayanan2017simple}.
We can view this ensemble as an approximation to a Bayesian model average, where the posterior is uniformly distributed over a simplicial complex.

Our paper is structured as follows: In Section \ref{sec: mode-volumes}, we introduce a method to discover multi-dimensional mode connecting simplexes in the neural network loss surface. In Section \ref{sec: volumes}, we show the existence of mode connecting volumes and provide a lower bound on the dimensionality of these volumes. Building on these insights, in 
Section \ref{sec: espro} we introduce ESPRO, a state-of-the-art approach to ensembling with neural networks, which efficiently averages over simplexes. In Section \ref{sec: uncertainty}, we show that ESPRO also provides well-calibrated representations of uncertainty.
We emphasize that ESPRO can be used as a simple drop-in replacement for deep ensembles, with improvements in accuracy and uncertainty representations. Code is available at \url{https://github.com/g-benton/loss-surface-simplexes}.

\section{Related Work}
The study of neural network loss surfaces has long been intertwined with an understanding of neural network generalization. 
\citet{hochreiter1997flat} argued that \emph{flat} minima provide better generalization, and proposed an optimization algorithm to find such solutions.
\citet{keskar2016large} and \citet{li2018visualizing} reinvigorated this argument by visualizing loss surfaces and studying the geometric properties of deep neural networks at their minima.  \citet{izmailov2018averaging} found that averaging SGD iterates with a modified learning rate finds flatter solutions that generalize better.
\citet{maddox2019simple} leveraged these insights in the context of Bayesian deep learning to form posteriors in flat regions of the loss landscape. Moreover, \citet{maddox2020rethinking} found many directions in parameter space that can be perturbed without changing the training or test loss.

\citet{freeman2016topology} demonstrated that single layer ReLU neural networks can be connected along a low loss curves.
\citet{garipov2018loss} and \citet{draxler2018essentially}   simultaneously demonstrated that it is possible to find low loss curves for ResNets and other deep networks.
\citet{skorokhodov2019loss} used multi-point optimization to parameterize wider varieties of shapes in loss surfaces, when visualizing the value of the loss, including exotic shapes such as cows.
\citet{czarnecki2019deep} then showed that low dimensional spaces of nearly constant loss theoretically exist in the loss surfaces of deep ReLU networks, but did not provide an algorithm to find these loss surfaces.

\citet{fort2019large} propose viewing the loss landscape as a series of potentially connected low-dimensional wedges in the much higher dimensional parameter space.
They then demonstrate that sets of optima can be connected via low-loss connectors that are generalizations of \citet{garipov2018loss}'s procedure. Our work generalizes these findings by discovering higher dimensional mode connecting volumes, which we then leverage for a highly efficient and practical ensembling procedure. 

Also appearing at the same conference as this work, \citet{wortsman2021learning} concurrently proposed a closely related technique to learning low dimensional neural network subspaces by extending the methods of \citet{fort2019deep} and \citet{garipov2018loss}.
\citet{wortsman2021learning} propose learning simplexes in parameter space with a regularization penalty to encourage diversity in weight space.

\section{Mode Connecting Volumes}
\label{sec: mode-volumes}

We now show how to generalize the procedure of \citet{garipov2018loss} to discover simplices of mode connecting \emph{volumes}, containing infinitely many mode connecting curves. In Section \ref{sec: mode-volumes}, we then show how to use our procedure to demonstrate the existence of these volumes in modern neural networks, revising our understanding about the structure of their loss landscapes. In Sections \ref{sec: espro} and \ref{sec: uncertainty} we show how to we can use these discoveries to build practical new methods which provide state of the art performance for both accuracy and uncertainty representation. We refer to our approach as SPRO (Simplicial Pointwise Random Optimization).

\subsection{Simplicial Complexes of Low Loss}

To find mode connecting volumes we seek \emph{simplexes} and \emph{simplicial complexes} of low loss.
Two primary reasons we seek simplexes of low loss are that (i) simplexes are defined by only a few points, and (ii) simplexes are easily sampled. The first point means that to define a mode connecting simplicial complex of low loss we need only find a small number of vertices to fully determine the simplexes in the complex. The second point means that we have easy access to the models contained within the simplex, leading to the practical simplex-based ensembling methods presented later in the paper.

We consider data $\mathcal{D}$, and training objective $\mathcal{L}$. We refer to $S_{(a_0, a_1, \dots, a_k)}$ as the $k$-simplex formed by vertices $a_0, a_1, \dots, a_k$, and $\text{V}(S_{(a_0, \dots, a_k)})$ as the volume of the simplex.\footnote{We use Cayley-Menger determinants to compute the volume of simplexes; for more information see Appendix \ref{app:simplexes}.}
Simplicial complexes are denoted $\mathcal{K}(S_{(a_0, a_1, \dots, a_{N_a})}, S_{(b_0, b_1, \dots, b_{N_b})}, \dots, S_{(m_0, m_1, \dots, m_{N_m})})$, and their volume is computed as the sum of the volume of their components. 
We use $w_j$ to denote \emph{modes}, or SGD training solutions, and $\theta_j$ to denote mode connecting points. For example, we could train two independent models to find parameter settings $w_0$ and $w_1$, and then find mode connecting point $\theta_0$ such that the path $w_0 \rightarrow \theta_0 \rightarrow w_1$ traversed low loss parameter settings as in \citet{fort2019large} and \citet{garipov2018loss}.

\subsection{Simplicial Complexes With SPRO}
\label{sec: mode-conn-methods}

To find a simplicial complex of low loss solutions, we first find a collection of modes $w_0, \dots, w_k$ through standard training. This procedure gives the trivial simplicial complex $\mathcal{K}(S_{(w_0)}, \dots, S_{(w_k)})$ (or $\mathcal{K}$), a complex containing $k$ disjoint $0$-simplexes.
With these modes we can then iteratively add connecting points, $\theta_j$, to join any number of the $0$-simplexes in the complex, and train the parameters in $\theta_j$ such that the loss within the simplicial complex, $\mathcal{K}$, remains low. The procedure to train these connecting $\theta_j$ forms the core of the SPRO algorithm, given here.

To gain intuition, we first consider some examples before presenting the full SPRO training procedure.
As we have discussed, we can take modes $w_0$ and $w_1$ and train $\theta_0$ to find a complex $\mathcal{K}(S_{(w_0, \theta_0)}, S_{(w_1, \theta_0)})$, which recovers a mode connecting path as in \citet{garipov2018loss}. Alternatively, we could connect $\theta_0$ with more than two modes and build the complex $\mathcal{K}(S_{(w_0, \theta_0)}, \dots, S_{(w_4, \theta_0)})$, connecting $5$ modes through a single point, similar to the $m$-tunnels presented in \citet{fort2019large}. SPRO can be taken further, however, and we could train (one at a time) a sequence of $\theta_j$'s to find the complex $\mathcal{K}(S_{(w_0, \theta_0, \theta_1, \theta_2)}, S_{(w_1, \theta_0, \theta_1, \theta_2)}, S_{(w_2, \theta_0, \theta_1, \theta_2)})$, describing a multi-dimensional volume that simultaneously connects $3$ modes through $3$ shared points.

We aim to train the $\theta_j$'s in $\mathcal{K}$ such that \textcolor{RedOrange}{the expected loss for models in the simplicial complex is low} and \textcolor{NavyBlue}{the volume of the simplicial complex is as large as possible}. 
That is, as we train the $j^{th}$ connecting point, $\theta_j$, we wish to minimize $\textcolor{RedOrange}{\mathbb{E}_{\phi \sim \mathcal{K}}\mathcal{L}(\mathcal{D}, \phi)}$ while maximizing $\textcolor{NavyBlue}{\text{V}(\mathcal{K})}$, using $\phi \sim \mathcal{K}$ to indicate $\phi$ follows a uniform distribution over the simplicial complex $\mathcal{K}$.

Following \citet{garipov2018loss}, we use $H$ parameter vectors randomly sampled from the simplex, $\phi_{h=1}^{H} \sim \mathcal{K}$, to compute $\textcolor{RedOrange}{\frac{1}{H}\sum_{h=1}^{H}\mathcal{L}(\mathcal{D}, \phi_h)}$ as an estimate of $\textcolor{RedOrange}{\mathbb{E}_{\phi \sim \mathcal{K}}\mathcal{L}(\mathcal{D}, \phi)}$.\footnote{We discuss the exact method for sampling, and the implications on bias in the loss estimate in Appendix \ref{app:simplexes}.} 
In practice we only need a small number of samples, $H$, and for all experiments use $H=5$ to balance between avoiding significant slowdowns in the loss function and ensuring we have reasonable estimates of the loss over the simplex.
Using this estimate we train $\theta_j$ by minimizing the regularized loss,
\begin{equation}
	\label{eqn:reg-loss}
	\begin{aligned}
		\mathcal{L}_{reg}(\mathcal{K}) &= \textcolor{RedOrange}{\frac{1}{H}\sum_{\phi_h \sim \mathcal{K}}\mathcal{L}(\mathcal{D}, \phi_h)} -
		\textcolor{NavyBlue}{\lambda_j \log(\text{V}(\mathcal{K}))}.
	\end{aligned}
\end{equation}
The regularization penalty $\lambda_j$ balances the objective between seeking a smaller simplicial complex that contains strictly low loss parameter settings (small $\lambda_j$), and a larger complex that that may contain less accurate solutions but encompasses more volume in parameter space (large $\lambda_j$). In general only a small amount of regularization is needed, and results are not sensitive to the choice of $\lambda_j$. In Section \ref{sec: espro} we explain how to adapt Eq. \ref{eqn:reg-loss} to train simplexes of low loss using single independetly trained models.... We provide details about how we choose $\lambda_j$ in Appendix~\ref{app: methods-details}.

\section{Volume Finding Experiments}
\label{sec: volumes}

In this section, we find volumes of low loss in a variety of settings. First, we show that the mode finding procedure of \citet{garipov2018loss} can be extended to find distributions of modes.
Then, we explore mode connecting simplicial complexes of low loss in a variety of settings, and finally provide an empirical upper bound on the dimensionality of the mode connecting spaces.

\paragraph{Loss Surface Plots.} Throughout this section and the remainder of the paper we display two-dimensional visualizations of loss surfaces of neural networks. These plots represent the loss within the plane defined by the three points (representing parameter vectors) in each plot. More specifically, if the three points in question are, e.g., $w_0$, $w_1$, and $w_2$ then we define $c = \frac{1}{3}\sum_{i=0}^{2}w_i$ as the center of the points and use Gram-Schmidt to construct $u$ and $v$, an orthonormal basis for the plane defined by the points. With the center and the basis chosen, we can sample the loss at parameter vectors of the form $w = c + r_u u + r_v v$ where $r_u$ and $r_v$ range from $-R$ to $R$, a range parameter chosen such that all the points are within the surface with a reasonable boundary. 

\subsection{Volumes of Connecting Modes}
\label{sec: swag_modes}

In Bayesian deep learning, we wish to form a predictive distribution through a posterior weighted Bayesian model average:
\begin{align}
p(y|x, \mathcal{D}) = \int p(y|w, x) p(w|\mathcal{D}) dw \,,
\label{eqn: bma}
\end{align}
where $y$ is an output (e.g., a class label), $x$ is an input (e.g., an image), $\mathcal{D}$ is the data, and $w$ are the neural network weights. This integral is challenging to compute due to the complex structure of the posterior $p(w|\mathcal{D})$. 

To help address this challenge, we can instead approximate the Bayesian model average in a subspace that contains many good solutions, as in \citet{izmailov2020subspace}. Here, we generalize the 
mode connecting procedure of 
\citet{garipov2018loss} to perform inference over subspaces that contain \emph{volumes} of mode connecting curves.

In \citet{garipov2018loss}, a mode connecting curve is defined by its parameters $\theta$. 
Treating the objective used to find $\theta$ in \citet{garipov2018loss}, $l(\theta)$, as a likelihood, we infer an approximate Gaussian posterior $q(\theta|\mathcal{D})$ using the SWAG procedure of \citet{maddox2019simple}, which induces a distribution over mode connecting curves. Each sample from $q(\theta|D)$ provides a mode connecting curve, which itself contains a space of complementary solutions.

\begin{figure}[h!]
    \centering
        \includegraphics[width=0.7\linewidth,trim={0 7.5cm 0 7.5cm},clip]{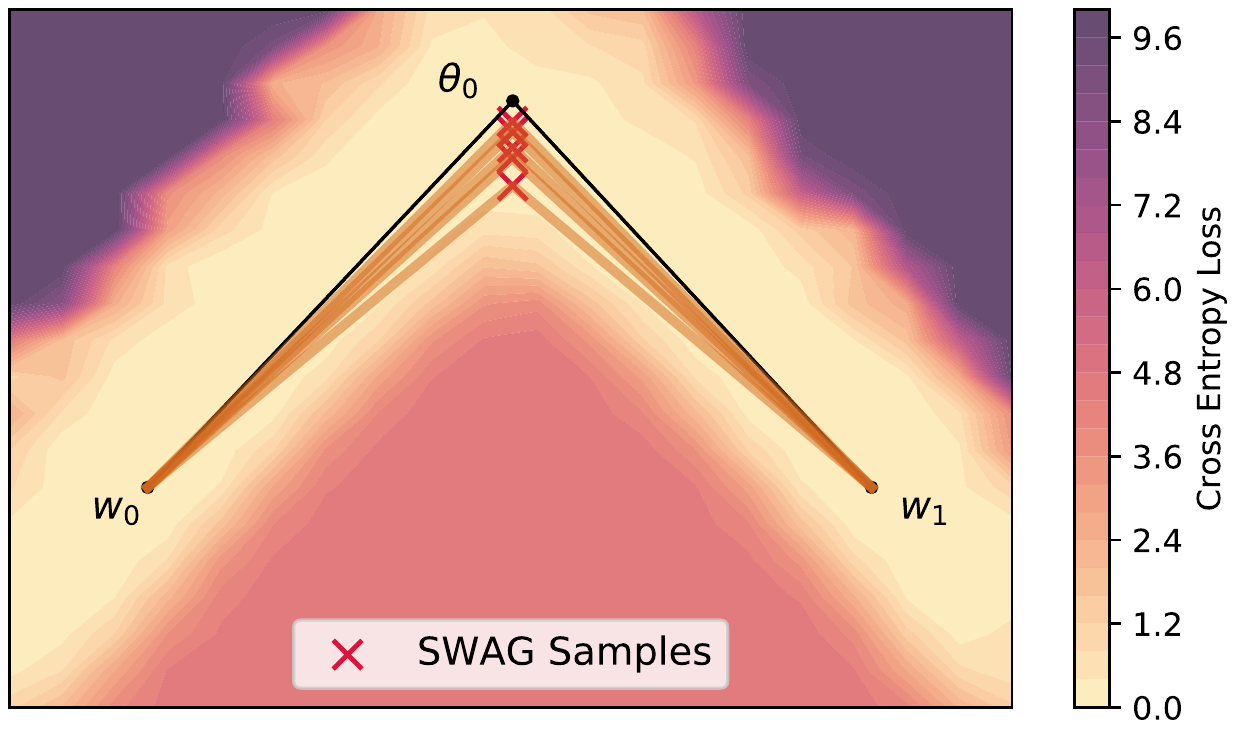}
    \caption{A loss surface in the basis spanned by the defining points of a connecting curve, $w_0, w_1, \theta_0$. Using SWAG, we form a posterior distribution over mode connecting curves, representing a volume of low loss explanations for the data.}
    \label{fig: swag_mode_connectors}
\end{figure}

In Figure~\ref{fig: swag_mode_connectors}, we see that it is possible to move between different values of $\theta$ without leaving a region of low loss.
We show samples from the SWAG posterior, projected into the plane formed by the endpoints of the curves, $w_0$ and $w_1$, and a  mode connecting point $\theta_0$. We show the induced connecting paths from SWAG samples with orange lines.
All samples from the SWAG posterior lie in the region of low loss, as do the sampled connecting paths, indicating that there is indeed an entire volume of connected low loss solutions induced by the SWAG posterior over $\theta$.
We provide training details in the Appendix \ref{app:swag_connectors}.

\subsection{Simplicial Complex Mode Connectivity}\label{sec: mode-connectivity}

The results of Section \ref{sec: swag_modes} suggest that modes might be connected by \emph{multi-dimensional} paths.
SPRO represents a natural generalization of the idea of learning a distribution over connecting paths.
By construction, if we use SPRO to find the simplicial complex $\mathcal{K}(S_{(w_0, \theta_0, \dots, \theta_k)}, \dots, S_{(w_m, \theta_0, \dots, \theta_k)})$ we have found a whole \emph{space} of suitable vertices to connect the modes $w_0, \cdots, w_m$. 
Any $\theta$ sampled from the $k$-simplex $S_{(\theta_0, \dots,\theta_k)}$ will induce a low-loss connecting path between any two vertices in the complex. 

To demonstrate that SPRO finds volumes of low loss, we trained a simplicial complex using SPRO, $\mathcal{K}(S_{(w_0, \theta_0, \theta_1, \theta_2)}, S_{(w_1, \theta_0, \theta_1, \theta_2)}),$ forming two simplexes containing three connecting vertices $\theta_0, \theta_1, \theta_2$ between the two fixed points, $w_0$ and $w_1,$ which are pre-trained models.

\begin{figure}[t!]
\centering 
\includegraphics[width=\linewidth,trim={0 7cm 0 7cm},clip]{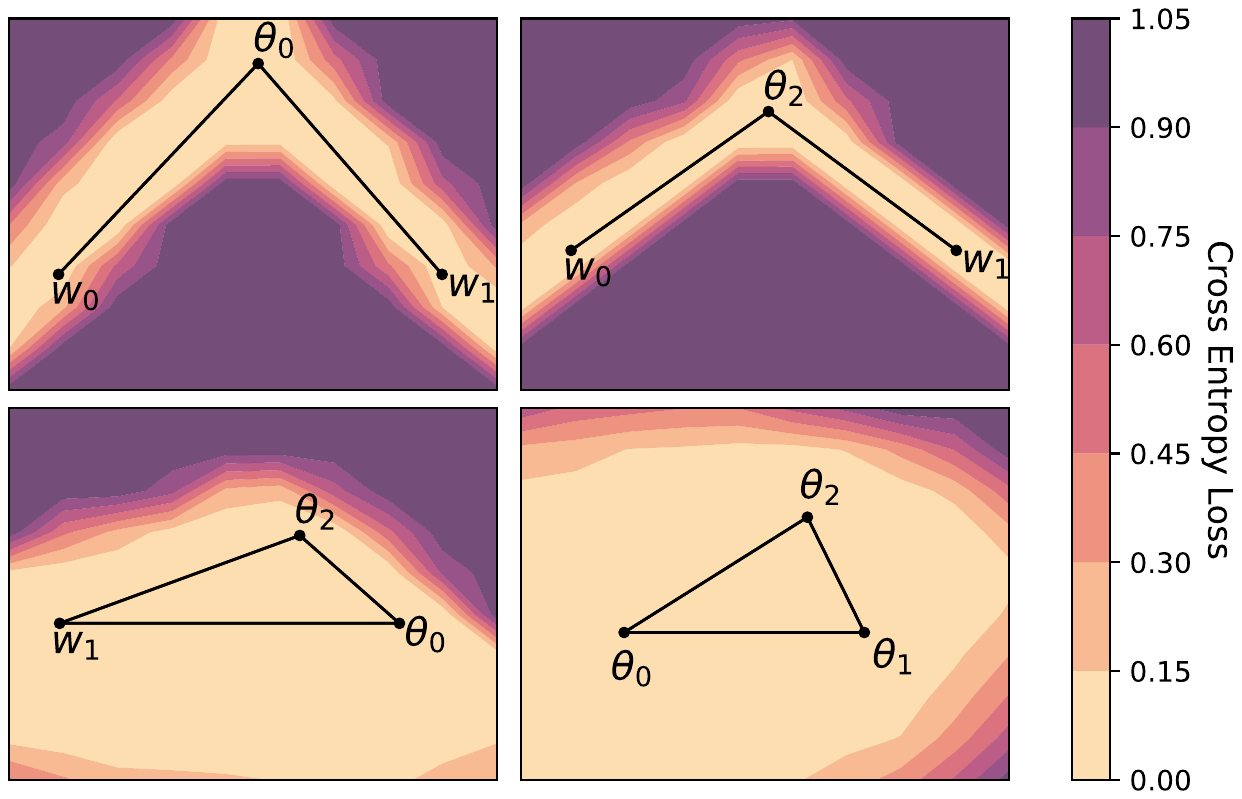}
\caption{
Loss surfaces for planes intersecting a mode connecting simplicial complex $\mathcal{K}(S_{(w_0, \theta_0, \theta_1, \theta_2)}, S_{(w_1, \theta_0, \theta_1, \theta_2)})$ trained on CIFAR-$10$ using a VGG-$16$ network. \textbf{Top:} along any $w_0 \rightarrow \theta_j \rightarrow w_1$ path we recover a standard mode connecting path. \textbf{Bottom Left:} a face of one of the simplexes that contains one of the independently trained modes. We see that as we travel away from $w_1$ along any path within the simplex we retain low train loss. \textbf{Bottom Right:} the simplex defined by the three mode connecting points. Any point sampled from within this simplex defines a low-loss mode connecting path between $w_0$ and $w_1$.
}
\label{fig:complex-connector}
\end{figure}

Figure \ref{fig:complex-connector} shows loss surface visualizations of this simplicial complex in the parameter space of a VGG-$16$ network trained on CIFAR-$10$.  We see that this complex contains not only standard mode connecting paths, but also volumes of low loss that connect modes. Figure \ref{fig:complex-connector} is a straightforward representation of how the loss landscape of large neural networks should be understood as suggested in Figure \ref{fig:loss_surface_theoretical}; not only are all training solutions connected by paths of low loss, they are points on the same \emph{multi-dimensional manifold} of low loss. In the bottom right panel of Figure~\ref{fig:complex-connector}, every point in the simplex corresponds to a different mode connecting curve.

\begin{figure}[t!]
    \begin{subfigure}[h]{0.23\textwidth}
        \centering
        \includegraphics[width=\textwidth]{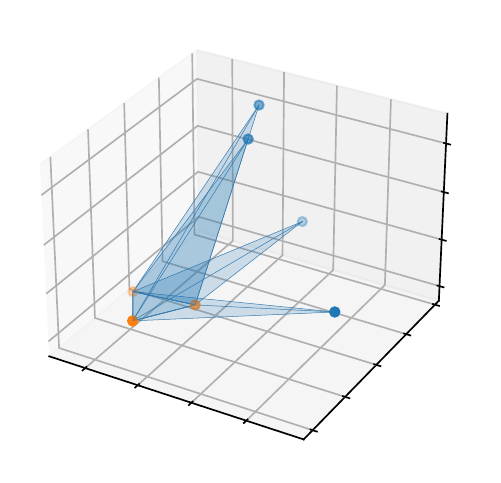}
        \caption{$4$ modes, $3$ connectors.}
        \label{fig: simp-conn-ex_left}
    \end{subfigure}
    \hfill
    \begin{subfigure}[h]{0.23\textwidth}
        \includegraphics[width=\textwidth]{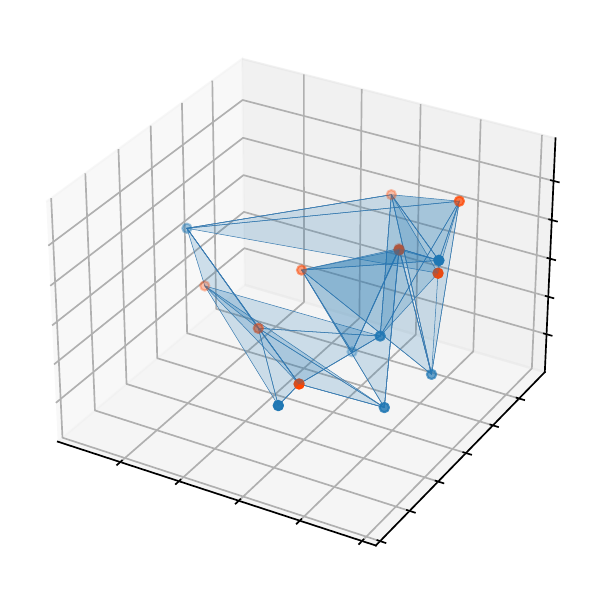}
        \caption{$7$ modes, $9$ connectors.}
        \label{fig: simp-conn-ex_right}
    \end{subfigure}
    \caption{\textbf{(a,b)} Three dimensional projections of mode connecting simplicial complexes with training modes shown in blue and connectors in orange. Blue shaded regions represent regions of low loss found via SPRO. \textbf{(a)} $4$ modes and $3$ connecting points found with a VGG-$16$ network on CIFAR-$100$. \textbf{(b)} $7$ modes and a total of $9$ connecting points found with a VGG-$16$ network on CIFAR-$10$.}
    \label{fig: simp-conn-ex}
\end{figure}

In Figure~\ref{fig: simp-conn-ex}, we show there exist manifolds of low loss that are vastly more intricate and high dimensional than a simple composition of $3$-simplexes connecting two modes.
In Figure \ref{fig: simp-conn-ex_left}, we connect $4$ modes using $3$ connecting points so that we have four different simplexes formed between the modes of low loss
for VGG-$16$ networks \citep{simonyan2014very} on CIFAR-$100$. 
The structure becomes considerably more intricate as we expand the amount of modes used; Figure \ref{fig: simp-conn-ex_right} uses $7$ modes with $9$ connecting points, forming $12$ inter-connected simplexes. Note that in this case not all modes are in shared simplexes with all connecting points.
These results clearly demonstrate that SPRO is capable of finding intricate and multi-dimensional structure within the loss surface. As a broader takeaway, \emph{any} mode we find through standard training is a single point within a large and high dimensional structure of loss, as shown in the rightmost representation in Figure \ref{fig:loss_surface_theoretical}. We consider the accuracy of ensembles found via these mode connecting simplexes in Appendix \ref{app:mode_espro}. In Section~\ref{sec: ensembles} we consider a particularly practical approach to ensembling with SPRO.

\begin{figure}[h!]
    \centering
    \includegraphics[height=0.49\textwidth,trim={6cm 0cm 6cm 0cm},clip,angle=90]{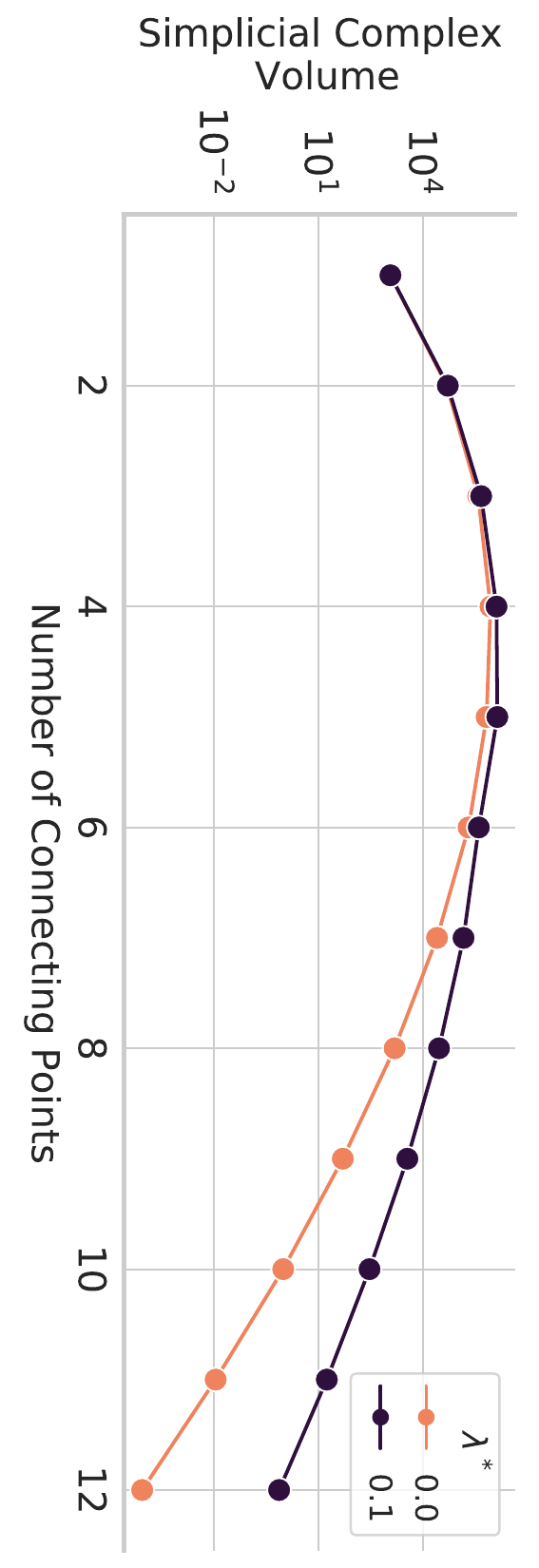}
    \caption{Volume of the simplicial complex as a function of the number of connectors for a VGG net on CIFAR-$10$ for two settings $\lambda$ of SPRO regularization. After $10$ connectors, the volume collapses, indicating that new points added to the simplicial complex are within the span of previously found vertices. The low-loss manifold must be at least $10$ dimensions in this instance.}
     \label{fig: conn-limit-vol}
\end{figure}

\subsection{Dimensionality of Loss Valleys}

We can estimate the highest dimensionality of the connecting space that SPRO can find, which provides a lower bound on the true dimensionality of these mode connecting subspaces for a given architecture and dataset.
To measure dimensionality, we take two pre-trained modes, $w_0$ and $w_1,$ and construct a connecting simplex with as many connecting points as possible, by finding the largest $k$ such that $\mathcal{K}(S_{(w_0, \theta_0, \dots, \theta_k)},S_{(w_1, \theta_0, \dots, \theta_k)})$ contains both low loss parameter settings and has non-zero volume. 
We could continue adding more degenerate points to the simplex; however, the resulting simplicial complex has no volume.

Figure \ref{fig: conn-limit-vol} shows the volume of a simplicial complex connecting two modes as a function of the number of connecting points, $k,$ for a VGG-$16$ network on CIFAR-$10$. To ensure these are indeed low-loss complexes, we sample $25$ models from each of these simplicial complexes and find that all sampled models achieve greater than $98\%$ accuracy on the train set.
We can continue adding new modes until we reach $k=11,$ when the volume collapses to approximately $10^{-4},$ from a maximum of $10^5.$ Thus the dimensionality of the manifold of low loss solutions for this architecture and dataset is at least $10$, as adding an eleventh point collapses the volume.

\section{ESPRO: Ensembling with SPRO} 
\label{sec: espro}

The ability to find large regions of low loss solutions has significant practical implications: we show how to use SPRO to efficiently create ensembles of models either within a single simplex or by connecting an entire simplicial complex.
We start by generalizing the methodology presented in Section \ref{sec: mode-conn-methods}, leading to a simplex based ensembling procedure, we call ESPRO (Ensembling SPRO). Crucially, our approach finds a low-loss simplex starting from only a \emph{single} SGD solution. We show that the different parameters in these simplexes gives rise to a diverse set of functions, which is crucial for ensembling performance.
Finally, we demonstrate that ESPRO outperforms state-of-the-art \emph{deep ensembles} \citep{lakshminarayanan2017simple}, both as a function of ensemble components and total computational budget. In Section~\ref{sec: uncertainty}, we show ESPRO also provides state-of-the-art results for uncertainty representation.

\subsection{Finding Simplexes from a Single Mode}

\begin{figure}[t!]
    \centering
    \includegraphics[width=\linewidth,trim={0 7cm 0 7cm},clip]{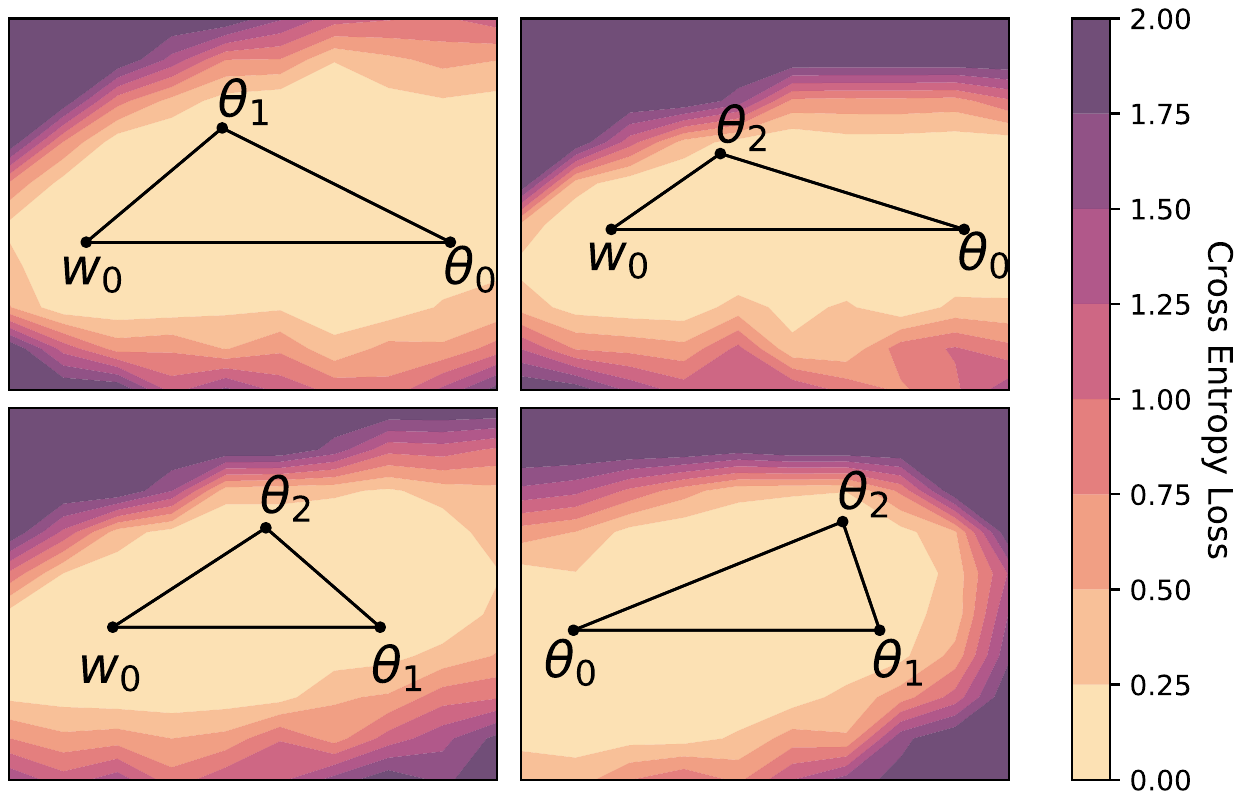}
    \caption{Loss surface visualizations of the faces of a sample ESPRO $3$-simplex for a VGG network trained on CIFAR-$100$. The ability to find a low-loss simplex starting from only a \emph{single} SGD solution, $w_0$, leads to an efficient ensembling procedure.}
    \label{fig:c100-espro-simplex}
\end{figure}

In Section \ref{sec: mode-conn-methods} we were concerned with finding a simplicial complex that connects multiple modes. We now describe how to adapt SPRO into a practical approach to ensembling by instead finding multiple simplexes of low loss, each --- crucially --- starting from a \emph{single} pre-trained SGD solution.

Simplexes contain a single mode, and take the form $S_{(w_j, \theta_{j,0}, \dots, \theta_{j,k})}$ where the $\theta_{j,k}$ is the $k^{th}$ vertex found with SPRO in a simplex where one of the vertices is mode $w_j$. We find SPRO simplexes one at a time, rather than as a complex. The associated loss function to find the $k^{th}$ vertex in association with mode $w_j$ is
\begin{equation}
	\label{eqn:espro-reg-loss}
	\begin{aligned}
		\mathcal{L}_{reg}(\mathcal{D}, &S_{(w_j, \theta_{j,0}, \dots, \theta_{j,k})}) = \textcolor{RedOrange}{\frac{1}{H}\sum_{\phi_h \sim S}\mathcal{L}(\mathcal{D}, \phi_h)} -\\
		&\textcolor{NavyBlue}{\lambda_i \log(\text{V}(S_{(w_j, \theta_{j,0}, \dots, \theta_{j,k})}))}.
	\end{aligned}
\end{equation}
For compactness we write $\phi_h \sim S$ to indicate $\phi_h$ is sampled uniformly at random from simplex $S_{(w_j, \theta_{j,0}, \dots, \theta_{j,k})}$.

We can think of this training procedure as extending out from the pre-trained mode $w_j$. First, in finding $\theta_{j,0}$ we find a line segment of low loss solutions, where one end of the line is $w_j$. Next, with $\theta_{j,0}$ fixed, we seek $\theta_{j, 1}$ such that the triangle formed by $w_j$, $\theta_{j,0}, \theta_{j,1}$ contains low loss solutions. 
We can continue adding vertices, constructing many dimensional simplexes.

With the resulting simplex $S_{(w_j, \theta_{j,0}, \dots, \theta_{j,k})}$, we can sample as many models from within the simplex as we need, and use them to form an ensemble.
Functionally, ensembles sampled from SPRO form an approximation to Bayesian marginalization over the model parameters where we assume a posterior that is uniform over the simplex.
We can define our prediction for a given input $x$ as,
\begin{equation}\label{eqn:ensemble-pred}
	\begin{aligned}
		\hat{y} &= \frac{1}{M}\sum_{\phi_m \sim S}f(x, \phi_m)\approx \int_{\phi_m \in S}f(x, \phi_h)d\phi_h,
	\end{aligned}
\end{equation}
where we write $S$ as shorthand for $S_{(w_j, \theta_{j,0}, \dots, \theta_{j,k})}.$
Specifically, the Bayesian model average and its approximation using approximate posteriors is 
\begin{align*}
    p(y^* | y, \mathcal{M}) &= \int p(y^* | \phi) p(\phi | y) d\phi \approx \int p(y^* | \phi) q(\phi | y) d\phi \\
    &\approx \frac{1}{M} \sum_{i=1}^M p(y^* | \phi_i); \quad \phi_i \sim q(\phi | y)
\end{align*}

\subsection{ESPRO: Ensembling over Multiple Independent Simplexes}\label{sec: method-ensemble}
We can significantly improve performance by ensembling from a simplicial complex containing multiple disjoint simplexes, which we refer to as ESPRO (Ensembling over SPRO simplexes).
To form such an ensemble, we take a collection of $j$ parameter vectors from independently trained models, $w_0, \dots, w_j$, and train a $k+1$-order simplex at each one using ESPRO.
This procedure defines the simplicial complex $\mathcal{K}(S_{(w_0, \dots, \theta_{0, k})}, \dots, S_{(w_j, \dots, \theta_{j, k})})$, which is composed of $j$ disjoint simplexes in parameter space. 
Predictions with ESPRO are generated as,
\begin{equation}\label{eqn:multi-ensemble-pred}
	\begin{aligned}
		\hat{y} &= \frac{1}{J}\sum_{\phi_j \sim \mathcal{K}}f(x, \phi_j)\approx \int_{\mathcal{K}}f(x, \phi_j)d\phi_j
	\end{aligned}
\end{equation}
where $\mathcal{K}$ is shorthand for $\mathcal{K}(S_{(w_0, \dots, \theta_{0, k})}, \dots, S_{(w_j, \dots, \theta_{j, k})})$.
ESPRO can be considered a mixture of simplexes (e.g. a simplicial complex) to approximate a multimodal posterior, towards a more accurate Bayesian model average. This observation is similar to how \citet{wilson2020bayesian} show that deep ensembles provide a compelling approximation to a Bayesian model average (BMA), and improve on deep ensembles through the MultiSWAG procedure, which uses a mixture of Gaussians approximation to the posterior for a higher fidelity BMA. ESPRO further improves the approximation to the BMA, by covering a larger region of the posterior corresponding to low loss solutions with functional variability. This perspective helps explain why ESPRO improves both accuracy and calibration, through a richer representation of epistemic uncertainty.

We verify the ability of ESPRO to find a simplex of low loss starting from a single mode in Figure \ref{fig:c100-espro-simplex}, which shows the loss surface in the planes defined by the faces of a $3$-simplex found in the parameter space of a VGG-$16$ network trained on CIFAR-$100$.
The ability to find these simplexes is core to forming ESPRO ensembles, as they only take a small number of epochs to find, typically less than $10\%$ the cost of training a model from scratch, and they contain diverse solutions that can be ensembled to improve model performance.
Notably, we can sweep out a volume of low loss in parameter space \emph{without} needing to first find multiple modes, in contrast to prior work on mode connectivity \citep{draxler2018essentially, garipov2018loss, fort2019large}.
We show additional results with image transformers \citep{dosovitskiy2020image} on CIFAR-$100$ in Appendix \ref{app:transformer_ls}, emphasizing that these simplexes are not specific to a particular architecture.

\subsection{SPRO and Functional Diversity}
\label{sec: two-spirals}

\begin{figure}[t!]
\centering
\begin{subfigure}[b]{0.23\textwidth}
    \includegraphics[width=\linewidth]{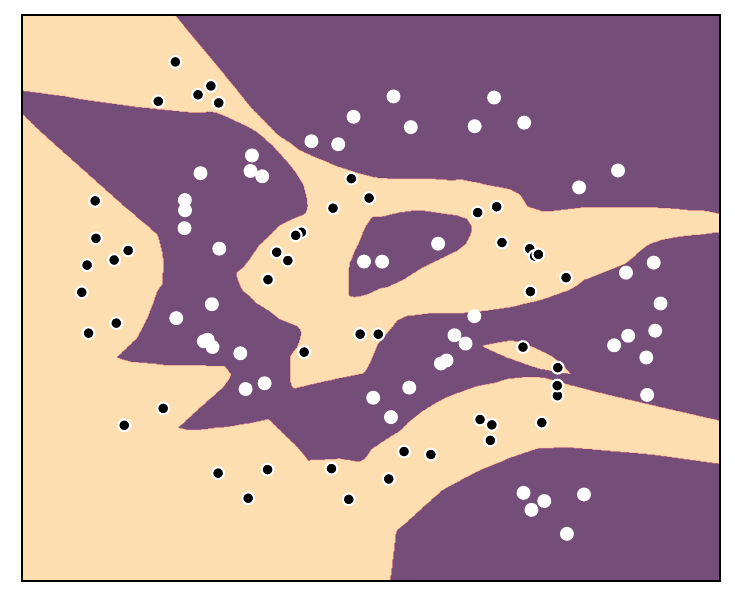}
\end{subfigure}
\begin{subfigure}[b]{0.23\textwidth}
    \includegraphics[width=\linewidth]{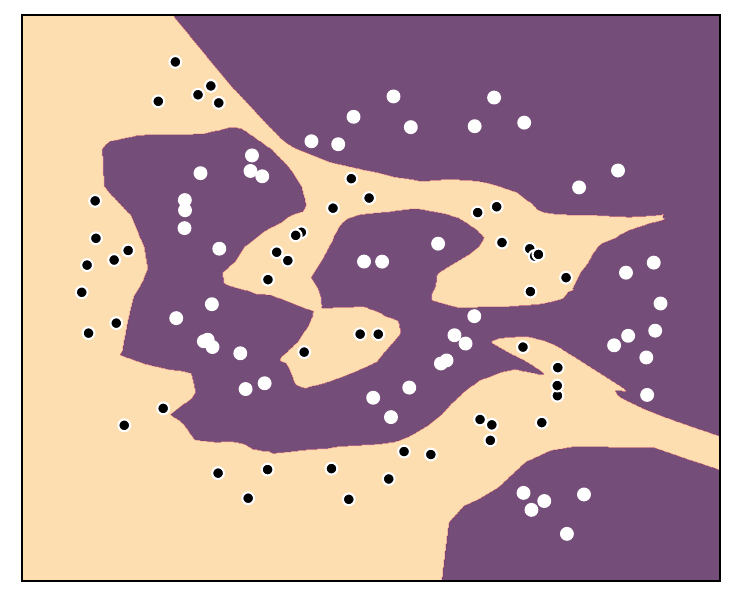}
\end{subfigure}
\caption{Functional diversity within a simplex. We show the decision boundaries for two classes, in the two spirals problem, with predictions in yellow and purple respectively.
Both plots are independent solution samples drawn from a $3$-simplex of an $8$-layer feed forward classifier and demonstrate that the simplexes have considerable functional diversity, as illustrated by different decision boundaries.
Significant differences are visible inside the data distribution (center of plots) and outside (around the edges).
}
\label{fig: two-spirals-ensemble}
\end{figure}
In practice we want to incorporate as many diverse high accuracy classifiers as possible when making predictions to gain the benefits of ensembling, such as improved accuracy and calibration. 
SPRO gives us a way to sample diverse models in \emph{parameter space}, and in this section we show, using a simple $2D$ dataset, that the parameter diversity found with SPRO is a reasonable proxy for the \emph{functional diversity} we actually seek.
\begin{figure*}[t!]
	\centering
	\includegraphics[height=\linewidth,trim={7.5cm 0 7.5cm 0},clip,angle=90]{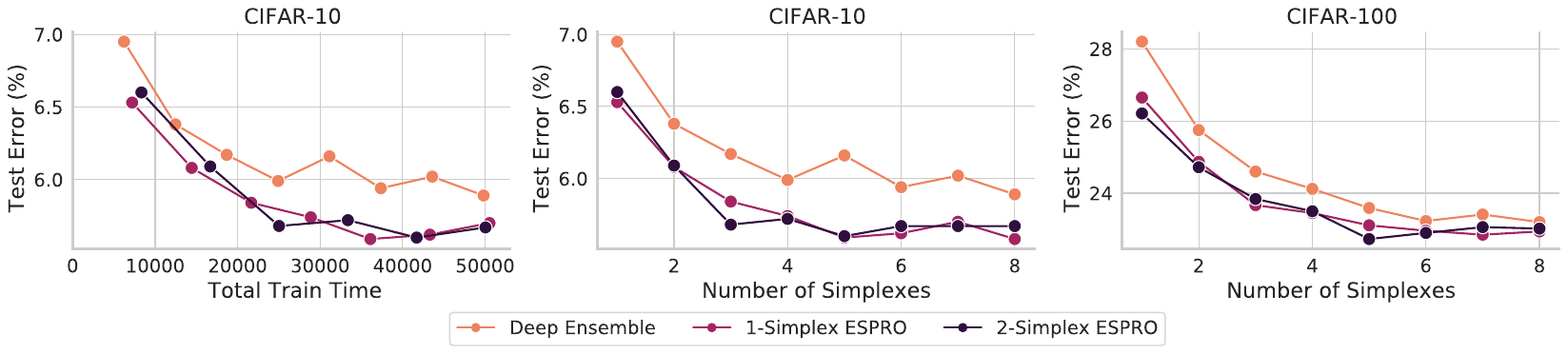}
	\caption{Performance of deep ensembles and ESPRO (with either a $1$-simplex, e.g. a line or a $2$-simplex, e.g. a triangle) using VGG-$16$ networks in terms of total train time and the number of simplexes (number of ensembles). \textbf{Left:} Test error as a function of total training budget on CIFAR-$10$. The number of components in the ensembles increases as curves move left to right. For any given training budget, ESPRO outperforms deep ensembles. \textbf{Center:} Test error as a function of the number of simplexes in the ensemble on CIFAR-$10$.
	A comparison of performance of ESPRO models on CIFAR-$10$ (\textbf{left}) and CIFAR-$100$ (\textbf{right}) of VGG-$16$ networks with various numbers of ensemble components along the $x$-axis, and various simplex orders indicated by color. For any fixed number of ensemble components we can outperform a standard deep ensemble using simplexes from ESPRO. Notably, expanding the number of vertices in a simplex takes \emph{only $10$ epochs of training} on CIFAR-$10$ compared to the $200$ epochs of training required to train a model from scratch. On CIFAR-$100$ adding a vertex to an ESPRO simplex takes just $20$ epochs of training compared to $300$ to train from scratch.}
	\label{fig: deep-ensembles}
\end{figure*}

To better understand how the simplexes interact with the functional form of the model, we consider an illustrative example on the two-spirals classification dataset presented in \citet{huang2019understanding}, in which predictions can be easily visualized. 
We find a $3$-simplex (a tetrahedron) in the parameter space of a simple $8$ layer deep feed forward classifier, and visualize the functional form of the model for both samples taken from within the simplex in parameter space. 
By examining the functional form of models sampled from simplexes in parameter space we can quickly see why ESPRO is beneficial. Figure \ref{fig: two-spirals-ensemble} shows individual models sampled from a single $3$-simplex in parameter space, corresponding to clear functional diversity. Models within the simplex all fit the training data nearly perfectly but do so in distinct ways, such that we can improve our final predictions by averaging over these models.

\subsection{Performance of Simplicial Complex Ensembles} \label{sec: ensembles}

Section \ref{sec: two-spirals} shows that we are able to discover simplexes in parameter space containing models that lead to diverse predictions, meaning that we can ensemble within a simplex and gain some of the benefits seen by deep ensembles \citep{lakshminarayanan2017simple}.
We use SPRO to train simplicial complexes containing a number of disjoint simplexes, and ensemble over these complexes to form predictions, using Eq. \ref{eqn:multi-ensemble-pred}.
We fix the number of samples taken from the ESPRO ensemble, $J$, to $25$ which provides the best trade off of accuracy vs test time compute cost.\footnote{We show the relationship between samples from the simplex and test error in Appendix \ref{app:error-v-sample}.}
For example, if we are training a deep ensemble of VGG-$16$ networks with $3$ ensemble components on CIFAR-$10$, we can form a deep ensemble to achieve an error rate of approximately $6.2$\%; however, by extending each base model to just a simple $2$-simplex ($3$ vertices) we can achieve an error rate of approximately $5.7\%$ --- an improvement of nearly $10\%$!

After finding a mode through standard training, a low order simplex can be found in just a small fraction of the time it takes to train a model from scratch. For a fixed training budget, we find that we can achieve a much lower error rate through training fewer overall ensemble components, but training low order simplexes (order $0$ to $2$) at each mode using ESPRO. 
Figure \ref{fig: deep-ensembles} shows a comparison of test error rate for ensembles of VGG-$16$ models over different numbers of ensemble components and simplex sizes on CIFAR-$10$ and CIFAR-$100$. For any fixed ensemble size, we can gain performance by using a ESPRO ensemble rather than a standard deep ensemble. 
Furthermore, training these ESPRO models is generally inexpensive; the models in Figure \ref{fig: deep-ensembles} are trained on CIFAR-$10$ for $200$ epochs and CIFAR-$100$ for $300$ epochs. Adding a vertex takes only an additional $10$ epochs of training on CIFAR-$10$, and $20$ epochs of training on CIFAR-$100$. We show the CIFAR-$100$ time-accuracy tradeoff in Appendix \ref{app:espro} finding a similar trend to CIFAR-$10$.

\begin{figure}[h!]
\centering
\includegraphics[width=0.8\linewidth,trim={0 7.5cm 0 7.5cm},clip]{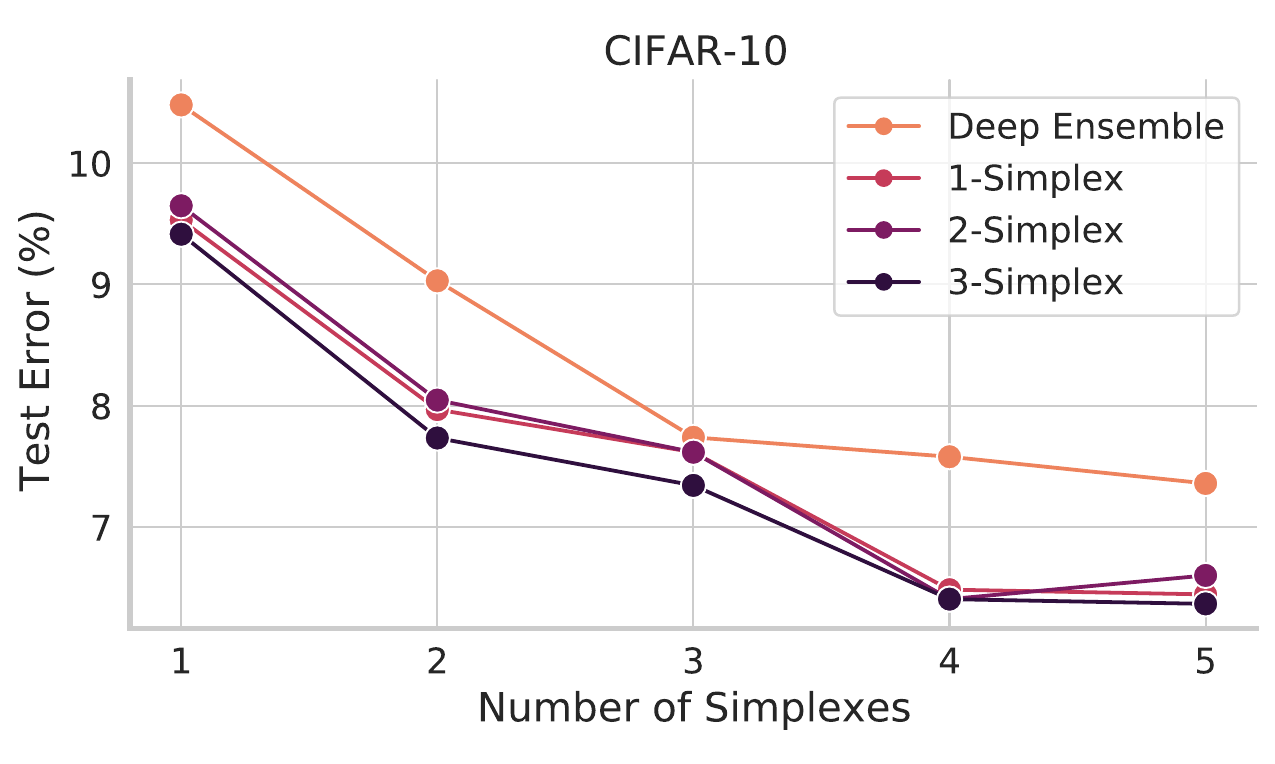}
\caption{Performance of deep ensembles and ESPRO ($1$, $2$, or $3$-simplex) using ResNet-$56$ models on CIFAR-$10$. The ResNet-$56$s follow the same trend as VGG networks: more ensemble components increases accuracy, ESPRO significantly outperforms deep ensembles, and adding further simplex vertices to each ESPRO component provides additional improvements.}
\label{fig:resnet_acc}
\end{figure}

Figure \ref{fig:resnet_acc} shows a comparison of test error for ensembles of ResNet-$56$ models over different ensemble and simplex sizes in Figure~\ref{fig: deep-ensembles}, providing more evidence for the general applicability of the ESPRO procedure. 
The main practical difference between  ResNet-$56$'s and the previous VGG networks is that the ResNet-$56$'s use BatchNorm. 
BatchNorm statistics need to be adjusted when we sample a model from within a simplex, leading to an additional cost at test time. 
To generate predictions, we use $100$ minibatches of train data to update the batch norm statistics before freezing the statistics and predicting on the test set.

\begin{figure}[h!]
\centering
\includegraphics[height=\linewidth,trim={2.75cm 0 2.75cm 0},clip,angle=90]{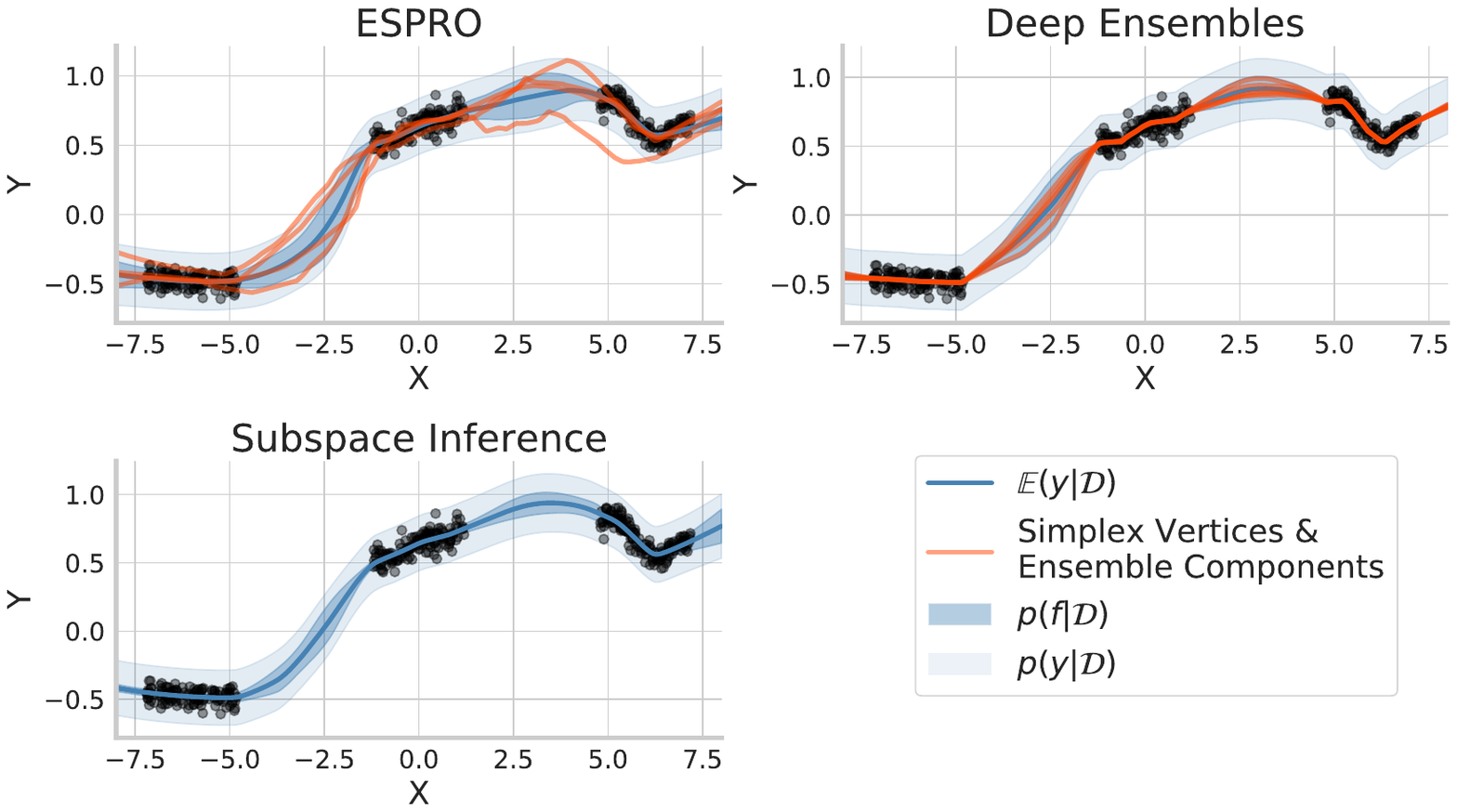}
\caption{Qualitative uncertainty plots of $p(f | \mathcal{D})$ on a  regression problem. We show both the $2\sigma$ confidence regions from $p(f | \mathcal{D})$ (the latent noise-free function) and $p(y | \mathcal{D}),$ which includes the observed noise of the data (aleatoric uncertainty). \textbf{Top Left:} ESPRO, colored lines are the vertices in the simplex. First two are fixed points in the simplex. \textbf{Top Right:} Deep ensembles, colored lines are individual models. \textbf{Bottom Left:} Curve subspaces.
ESPRO solutions produce functionally diverse solutions that have good in-between (between the data distribution) and extrapolation (outside of the data distribution) uncertainties; the ESPRO predictive distribution is broader and more realistic than deep ensembles and mode-connecting subspace inference, by containing a greater variety of high performing solutions.}
\label{fig:reg_qualitative}
\end{figure}

\begin{figure*}
    \begin{subfigure}{\linewidth}
       \includegraphics[height=\linewidth,trim={7.5cm 0cm 7.5cm 0cm},clip,angle=90]{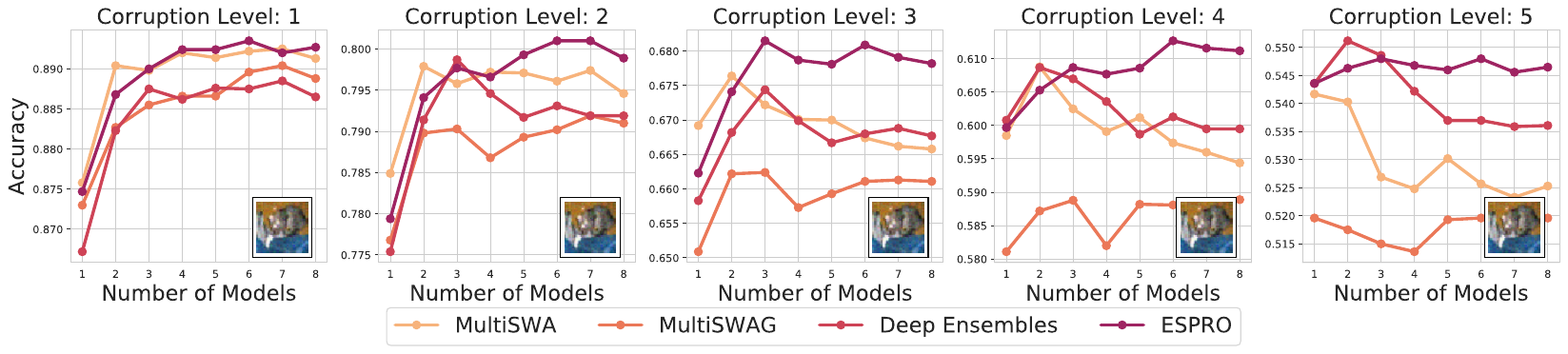}
        \caption{Accuracy for Gaussian noise corruption}
        \label{fig:corruption_acc}
    \end{subfigure}
    \begin{subfigure}{\linewidth}
        \includegraphics[height=\linewidth,trim={7.5cm 0cm 7.5cm 0cm},clip,angle=90]{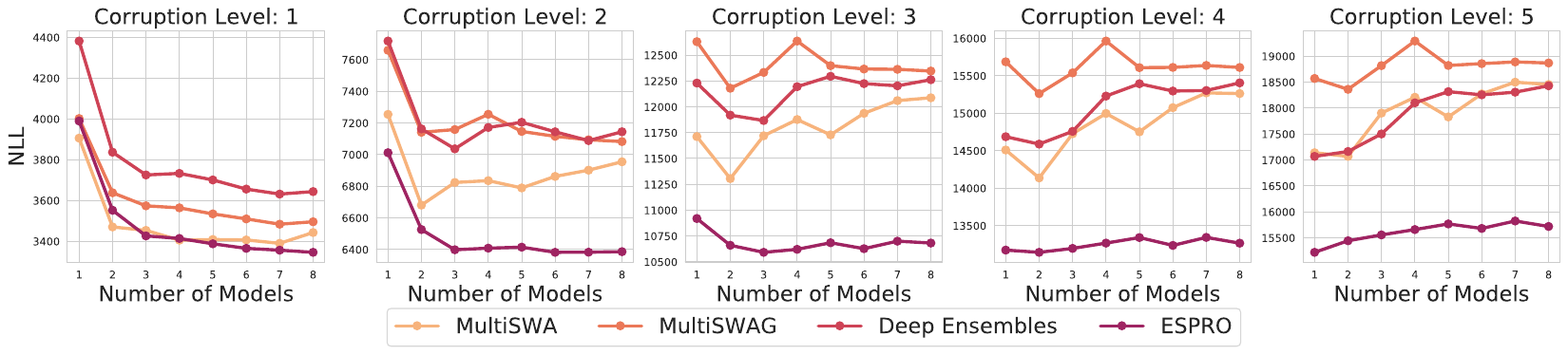}
        \caption{NLL for Gaussian noise corruption}
        \label{fig:corruption_nll}
    \end{subfigure}
    \centering
    \caption{\textbf{(a)} Accuracy for Gaussian blur corruption for MultiSWA, MultiSWAG, deep ensembles and ESPRO. \textbf{(b)} NLL under the same corruption. All models were originally significantly over-confident so we use temperature scaling \citep{guo2017calibration} to improve uncertainty; after temperature scaling ESPRO generally performs the best under varying levels of corruption.}
    \label{fig:acc_nll_ece}
\end{figure*}
\section{Uncertainty and Robustness} \label{sec: uncertainty}

We finish by investigating the uncertainty representation and robustness to dataset shift provided by ESPRO.
We show qualitative results on a regression problem, before studying corruptions of CIFAR-10, comparing to deep ensembles, MultiSWA, and the state-of-the-art Bayesian approach MultiSWAG \citep{wilson2020bayesian}.

\subsection{Qualitative Regression Experiments}
In general, a good representation of epistemic (model) uncertainty has the property that the uncertainty grows as we move away from the data.
Visualizing the growth in uncertainty is most straightforward in simple one-dimensional regression problems.

\citet{izmailov2020subspace} visualize one dimensional regression uncertainty by randomly initializing a two layer neural network, evaluating the neural network on three disjoint random inputs in one dimension: $(-7, -5)$, $(-1, 1),$ and $(5, 7),$ and adding noise of $\sigma^2 = 0.1$ to the net's outputs.
The task is to recover the true noiseless function, $f,$ given another randomly initialized two layer network, as well as to achieve reasonable confidence bands in the regions of missing data --- we used a Gaussian likelihood with fixed $\sigma^2 = 0.1$ to train the networks, modelling the noisy data $y$.
In Figure \ref{fig:reg_qualitative}, we show ESPRO (top left) which recovers good qualitative uncertainty bands on this task. 
We compare to deep ensembles (size $5$) (top right) and the state of the art subspace inference method of \citet{izmailov2020subspace} (bottom left), finding that ESPRO does a better job of recovering uncertainty about the latent function $f$ than either competing method, as shown by the $2\sigma$ confidence region about $p(f | \mathcal{D})$.
Indeed, after adding in the true noise, ESPRO complexes also do a better job of modelling the noisy responses, $y,$ measured by $p(y | \mathcal{D})$ than either approach.

\subsection{Uncertainty and Accuracy under Dataset Shift}

Modern neural networks are well known to be poorly calibrated and to result in overconfident predictions. 
Following \citet{ovadia2019can}, we consider classification accuracy, the negative log likelihood (NLL), and expected calibration error (ECE), to asses model performance under varying amounts of dataset shift, comparing to deep ensembles \citep{lakshminarayanan2017simple}, MultiSWA, and MultiSWAG \citep{wilson2020bayesian}, a state-of-the-art approach to Bayesian deep learning which generalizes deep ensembles.
In Figure~\ref{fig:corruption_acc}, we show results across all levels for the Gaussian noise corruption, where we see that ESPRO is most accurate across all levels.
For NLL we use temperature scaling \citep{guo2017calibration} on all methods to reduce the over-confidence and report the results in Figure~\ref{fig:corruption_nll}.  We see that ESPRO with temperature scaling outperforms all other methods for all corruption levels.
We show ECE and results across other types of dataset corruption in Appendix \ref{app:nll_calib}.

\section{Discussion}

We have shown that the loss landscapes for deep neural networks contain large multi-dimensional simplexes of low loss solutions. We proposed a simple approach, which we term SPRO, to discover these simplexes. We show how this geometric discovery can be leveraged to develop a highly practical approach to ensembling, which samples diverse and low loss solutions from the simplexes. Our approach improves upon state-of-the-art methods including deep ensembles and MultiSWAG, in accuracy and robustness. 
Overall, this paper provides a new understanding of how the loss landscapes in deep learning are structured: rather than isolated modes, or basins of attraction connected by thin tunnels, there are large multidimensional manifolds of connected solutions. 

This new understanding of neural network loss landscapes has many exciting practical implications and future directions. We have shown we can build state-of-the-art ensembling approaches from low less simplexes, which serve as a simple drop-in replacement for deep ensembles. In the future, one could build posterior approximations that cover these simplexes, while extending coverage to lower density points for a more exhaustive Bayesian model average. 
We could additionally build stochastic MCMC methods designed to navigate specifically in these subspaces of low loss but diverse solutions. These types of topological features in the loss landscape, which are very distinctive to neural networks, hold the keys to understanding generalization in deep learning.

\section*{Acknowledgements}
GWB, WJM, SL, AGW are supported by an Amazon Research Award, NSF I-DISRE 193471, NIH R01 DA048764-01A1, NSF IIS-1910266, and NSF 1922658 NRT-HDR:FUTURE Foundations, Translation, and Responsibility for Data Science. 
WJM was additionally supported by an NSF Graduate Research Fellowship under Grant No.  DGE-1839302.
SL was additionally supported by a DeepMind Fellowship.

\clearpage
\bibliography{refs}

\begin{thebibliography}{26}
\providecommand{\natexlab}[1]{#1}
\providecommand{\url}[1]{\texttt{#1}}
\expandafter\ifx\csname urlstyle\endcsname\relax
  \providecommand{\doi}[1]{doi: #1}\else
  \providecommand{\doi}{doi: \begingroup \urlstyle{rm}\Url}\fi

\bibitem[Choromanska et~al.(2015)Choromanska, Henaff, Mathieu, Arous, and
  LeCun]{choromanska2015loss}
Choromanska, A., Henaff, M., Mathieu, M., Arous, G.~B., and LeCun, Y.
\newblock The loss surfaces of multilayer networks.
\newblock In \emph{Artificial intelligence and statistics}, pp.\  192--204.
  PMLR, 2015.

\bibitem[Colins()]{simplexVolume}
Colins, K.~D.
\newblock Cayley-menger determinant.
\newblock From MathWorld--A Wolfram Web Resource, created by Eric W. Weisstein,
  https://mathworld.wolfram.com/Cayley-MengerDeterminant.html.

\bibitem[Czarnecki et~al.(2019)Czarnecki, Osindero, Pascanu, and
  Jaderberg]{czarnecki2019deep}
Czarnecki, W.~M., Osindero, S., Pascanu, R., and Jaderberg, M.
\newblock A deep neural network's loss surface contains every low-dimensional
  pattern.
\newblock \emph{arXiv preprint arXiv:1912.07559}, 2019.

\bibitem[Dosovitskiy et~al.(2021)Dosovitskiy, Beyer, Kolesnikov, Weissenborn,
  Zhai, Unterthiner, Dehghani, Minderer, Heigold, Gelly,
  et~al.]{dosovitskiy2020image}
Dosovitskiy, A., Beyer, L., Kolesnikov, A., Weissenborn, D., Zhai, X.,
  Unterthiner, T., Dehghani, M., Minderer, M., Heigold, G., Gelly, S., et~al.
\newblock An image is worth 16x16 words: Transformers for image recognition at
  scale.
\newblock In \emph{International Conference on Learning Representations},
  volume~9, 2021.
\newblock URL \url{arXiv:2010.11929}.

\bibitem[Draxler et~al.(2018)Draxler, Veschgini, Salmhofer, and
  Hamprecht]{draxler2018essentially}
Draxler, F., Veschgini, K., Salmhofer, M., and Hamprecht, F.
\newblock Essentially no barriers in neural network energy landscape.
\newblock In \emph{International Conference on Machine Learning}, pp.\
  1309--1318, 2018.

\bibitem[Fort \& Jastrzebski(2019)Fort and Jastrzebski]{fort2019large}
Fort, S. and Jastrzebski, S.
\newblock Large scale structure of neural network loss landscapes.
\newblock In \emph{Advances in Neural Information Processing Systems},
  volume~32, pp.\  6709--6717, 2019.

\bibitem[Fort et~al.(2019)Fort, Hu, and Lakshminarayanan]{fort2019deep}
Fort, S., Hu, H., and Lakshminarayanan, B.
\newblock Deep ensembles: A loss landscape perspective.
\newblock \emph{arXiv preprint arXiv:1912.02757}, 2019.

\bibitem[Freeman \& Bruna(2017)Freeman and Bruna]{freeman2016topology}
Freeman, C.~D. and Bruna, J.
\newblock Topology and geometry of half-rectified network optimization.
\newblock In \emph{International Conference on Learning Representations}, 2017.
\newblock URL \url{arXiv:1611.01540}.

\bibitem[Garipov et~al.(2018)Garipov, Izmailov, Podoprikhin, Vetrov, and
  Wilson]{garipov2018loss}
Garipov, T., Izmailov, P., Podoprikhin, D., Vetrov, D.~P., and Wilson, A.~G.
\newblock Loss surfaces, mode connectivity, and fast ensembling of dnns.
\newblock \emph{Advances in Neural Information Processing Systems},
  31:\penalty0 8789--8798, 2018.

\bibitem[Goodfellow et~al.(2015)Goodfellow, Vinyals, and
  Saxe]{goodfellow2014qualitatively}
Goodfellow, I.~J., Vinyals, O., and Saxe, A.~M.
\newblock Qualitatively characterizing neural network optimization problems.
\newblock In \emph{International Conference on Learning Representations},
  volume~3, 2015.
\newblock URL \url{arXiv:1412.6544}.

\bibitem[Guo et~al.(2017)Guo, Pleiss, Sun, and Weinberger]{guo2017calibration}
Guo, C., Pleiss, G., Sun, Y., and Weinberger, K.~Q.
\newblock On calibration of modern neural networks.
\newblock In \emph{International Conference on Machine Learning}, volume~70,
  pp.\  1321--1330. PMLR, 2017.

\bibitem[Hochreiter \& Schmidhuber(1997)Hochreiter and
  Schmidhuber]{hochreiter1997flat}
Hochreiter, S. and Schmidhuber, J.
\newblock Flat minima.
\newblock \emph{Neural Computation}, 9\penalty0 (1):\penalty0 1--42, 1997.

\bibitem[Huang et~al.(2019)Huang, Emam, Goldblum, Fowl, Terry, Huang, and
  Goldstein]{huang2019understanding}
Huang, W.~R., Emam, Z., Goldblum, M., Fowl, L., Terry, J.~K., Huang, F., and
  Goldstein, T.
\newblock Understanding generalization through visualizations.
\newblock \emph{arXiv preprint arXiv:1906.03291}, 2019.

\bibitem[Izmailov et~al.(2018)Izmailov, Podoprikhin, Garipov, Vetrov, and
  Wilson]{izmailov2018averaging}
Izmailov, P., Podoprikhin, D., Garipov, T., Vetrov, D., and Wilson, A.~G.
\newblock Averaging weights leads to wider optima and better generalization.
\newblock In \emph{Uncertainty in Artificial Intelligence}, 2018.
\newblock URL \url{arXiv:1803.05407}.

\bibitem[Izmailov et~al.(2019)Izmailov, Maddox, Kirichenko, Garipov, Vetrov,
  and Wilson]{izmailov2020subspace}
Izmailov, P., Maddox, W.~J., Kirichenko, P., Garipov, T., Vetrov, D., and
  Wilson, A.~G.
\newblock Subspace inference for bayesian deep learning.
\newblock In \emph{Uncertainty in Artificial Intelligence}, pp.\  1169--1179.
  PMLR, 2019.

\bibitem[Keskar et~al.(2017)Keskar, Mudigere, Nocedal, Smelyanskiy, and
  Tang]{keskar2016large}
Keskar, N.~S., Mudigere, D., Nocedal, J., Smelyanskiy, M., and Tang, P. T.~P.
\newblock On large-batch training for deep learning: Generalization gap and
  sharp minima.
\newblock In \emph{International Conference on Learning Representations}, 2017.
\newblock URL \url{arXiv:1609.04836}.

\bibitem[Lakshminarayanan et~al.(2017)Lakshminarayanan, Pritzel, and
  Blundell]{lakshminarayanan2017simple}
Lakshminarayanan, B., Pritzel, A., and Blundell, C.
\newblock Simple and scalable predictive uncertainty estimation using deep
  ensembles.
\newblock \emph{Advances in neural information processing systems},
  30:\penalty0 6402--6413, 2017.

\bibitem[LeCun et~al.(1998)LeCun, Bottou, Bengio, and
  Haffner]{lecun1998gradient}
LeCun, Y., Bottou, L., Bengio, Y., and Haffner, P.
\newblock Gradient-based learning applied to document recognition.
\newblock \emph{Proceedings of the IEEE}, 86\penalty0 (11):\penalty0
  2278--2324, 1998.

\bibitem[Li et~al.(2018)Li, Xu, Taylor, Studer, and
  Goldstein]{li2018visualizing}
Li, H., Xu, Z., Taylor, G., Studer, C., and Goldstein, T.
\newblock Visualizing the loss landscape of neural nets.
\newblock In \emph{Advances in neural information processing systems}, pp.\
  6389--6399, 2018.

\bibitem[Maddox et~al.(2019)Maddox, Izmailov, Garipov, Vetrov, and
  Wilson]{maddox2019simple}
Maddox, W.~J., Izmailov, P., Garipov, T., Vetrov, D.~P., and Wilson, A.~G.
\newblock A simple baseline for bayesian uncertainty in deep learning.
\newblock \emph{Advances in Neural Information Processing Systems},
  32:\penalty0 13153--13164, 2019.

\bibitem[Maddox et~al.(2020)Maddox, Benton, and Wilson]{maddox2020rethinking}
Maddox, W.~J., Benton, G., and Wilson, A.~G.
\newblock Rethinking parameter counting in deep models: Effective
  dimensionality revisited.
\newblock \emph{arXiv preprint arXiv:2003.02139}, 2020.

\bibitem[Ovadia et~al.(2019)Ovadia, Fertig, Ren, Nado, Sculley, Nowozin,
  Dillon, Lakshminarayanan, and Snoek]{ovadia2019can}
Ovadia, Y., Fertig, E., Ren, J., Nado, Z., Sculley, D., Nowozin, S., Dillon,
  J., Lakshminarayanan, B., and Snoek, J.
\newblock Can you trust your model's uncertainty? evaluating predictive
  uncertainty under dataset shift.
\newblock In \emph{Advances in Neural Information Processing Systems},
  volume~32, pp.\  13991--14002, 2019.

\bibitem[Simonyan \& Zisserman(2015)Simonyan and Zisserman]{simonyan2014very}
Simonyan, K. and Zisserman, A.
\newblock Very deep convolutional networks for large-scale image recognition.
\newblock In \emph{International Conference on Learning Representations},
  volume~3, 2015.
\newblock URL \url{arXiv:1409.1556}.

\bibitem[Skorokhodov \& Burtsev(2019)Skorokhodov and
  Burtsev]{skorokhodov2019loss}
Skorokhodov, I. and Burtsev, M.
\newblock Loss landscape sightseeing with multi-point optimization.
\newblock \emph{arXiv preprint arXiv:1910.03867}, 2019.

\bibitem[Wilson \& Izmailov(2020)Wilson and Izmailov]{wilson2020bayesian}
Wilson, A.~G. and Izmailov, P.
\newblock Bayesian deep learning and a probabilistic perspective of
  generalization.
\newblock In \emph{Advances in Neural Information Processing Systems},
  volume~33, 2020.

\bibitem[Wortsman et~al.(2021)Wortsman, Horton, Guestrin, Farhadi, and
  Rastegari]{wortsman2021learning}
Wortsman, M., Horton, M., Guestrin, C., Farhadi, A., and Rastegari, M.
\newblock Learning neural network subspaces.
\newblock In \emph{Internatinal Conference on Machine Learning}, volume 139.
  PMLR, 2021.
\newblock URL \url{arXiv:2102.10472}.

\end{thebibliography}
\bibliographystyle{icml2021}

\renewcommand\thefigure{A.\arabic{figure}}
\renewcommand\theequation{A.\arabic{equation}}   
\renewcommand{\thesection}{\Alph{section}}
\setcounter{figure}{0}
\setcounter{equation}{0}

\onecolumn
\appendix
\icmltitle{Appendix for \\
Loss Surface Simplexes for Mode Connecting Volumes and Fast Ensembling}

\section*{Outline}
The Appendix is outlined as follows:

In Appendix \ref{app:simplexes}, we give a more detailed description of our methods, focusing first on computing the simplex volume and sampling from the simplexes, then describe vertex initialization and regularization, giving training details, and finally describing the training procedure for multi-dimensional mode connectors.

In Appendix \ref{app: methods-details}, we describe several more results on volume and ensembling, particularly on the number of samples required for good performance with SPRO and ESPRO.

Finally, in Appendix \ref{app:reg_uncertainty}, we plot the results of a larger suite of corruptions on CIFAR-$10$ for ESPRO, deep ensembles, and MultiSWAG.

\begin{figure}[H]
    \centering
    \includegraphics[width=\linewidth]{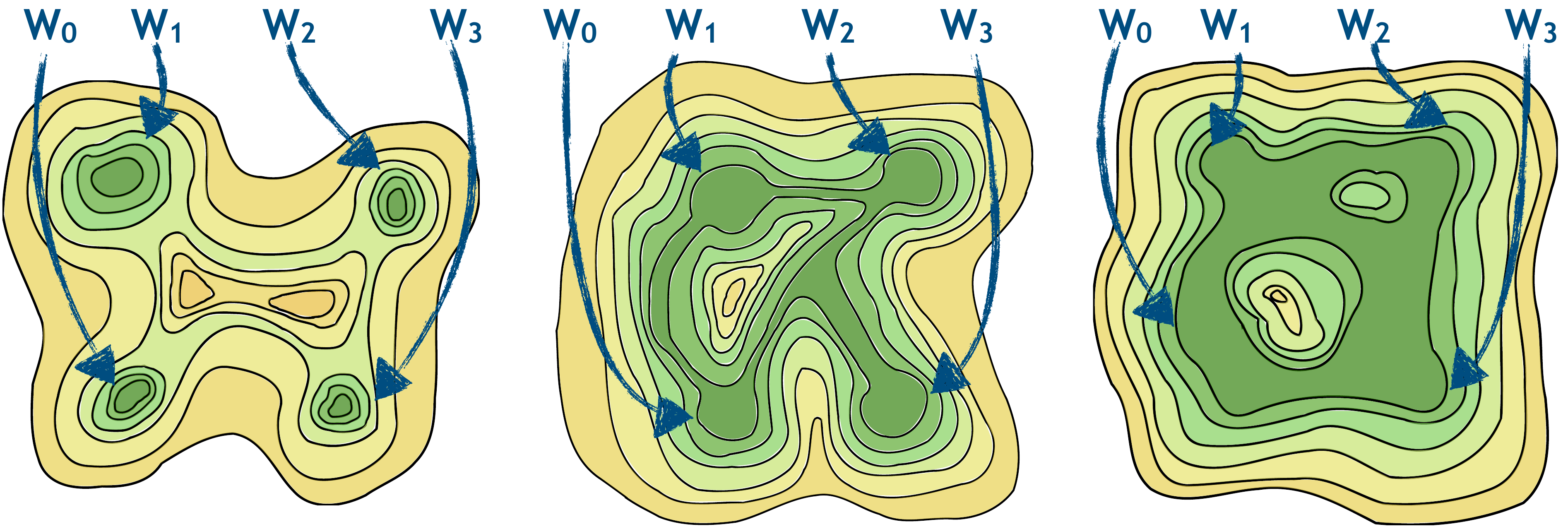}
    \caption{A simplified version of the progressive understanding of the loss landscape of neural networks. \textbf{Left:} The traditional view in which low loss modes are disconnected in parameter space. \textbf{Center:} The updated understanding provided by works such as \citet{draxler2018essentially}, \citet{fort2019large}, and \citet{garipov2018loss}, in which modes are connected along thin paths or tunnels. \textbf{Right:} The view we present in this work: independently trained models converge to points on the same \emph{volume} of low loss.}
    \label{fig: example-2D}
\end{figure}

\section{Extended Methodology} \label{app:extended_methods}

First, we present a two dimensional version of the schematic in Figure \ref{fig:loss_surface_theoretical} in \ref{fig: example-2D}, which explains the same progressive illustration, but in two dimensions. 
\subsection{Simplex Volume and Sampling} \label{app:simplexes}

We employ simplexes in the loss surface for two reasons primarily:
\begin{itemize}
    \item sampling uniformly from within a simplex is straightforward, meaning we can estimate the expected loss within any found simplexes easily,
    \item computing the Volume of a simplex is efficient, allowing for regularization encouraging high-Volume simplexes.
\end{itemize}

\paragraph{Sampling from Simplexes:} 
Sampling from the standard simplex is just a specific case of sampling from a Dirichlet distribution with concentration parameters all equal to $1$. 
The standard $n$-simplex is a simplex is a simplex formed by the vectors $\mathbf{v}_0, \dots, \mathbf{v}_n$ such that the $\mathbf{v}_i$'s are the standard unit vectors. 
Therefore, to draw samples from a standard $n$-simplex in a $d$ dimensional space with vertices $\mathbf{v}_0, \dots, \mathbf{v}_n$, we follow the same procedure to sample from a Dirichlet distribution. 

To sample vector $\mathbf{x} = [x_0, \dots, x_d]^T$ we first draw $y_0, \dots, y_n \overset{\text{i.i.d.}}{\sim} \textrm{Exp}(1)$, then set $\tilde{y}_i = \frac{y_i}{\sum_{j=1}^{d} y_j}$. Finally, $\mathbf{x} = \sum_{i=1}^{n}\tilde{y}_i \mathbf{v}_i$.

While this method is sufficient for simulating vectors uniformly at random from the \emph{standard} simplex, there is no guarantee that such a sampling method produces uniform samples from an arbitrary simplex, and thus samples of the loss over the simplex that we use in Equation \ref{eqn:reg-loss} may not be an unbiased estimate of the expected loss over the simplex. 
Practically, we do not find this to be an issue, and are still able to recover low loss simplexes with this approach. 

\begin{figure}
    \centering
    \includegraphics[width=\linewidth]{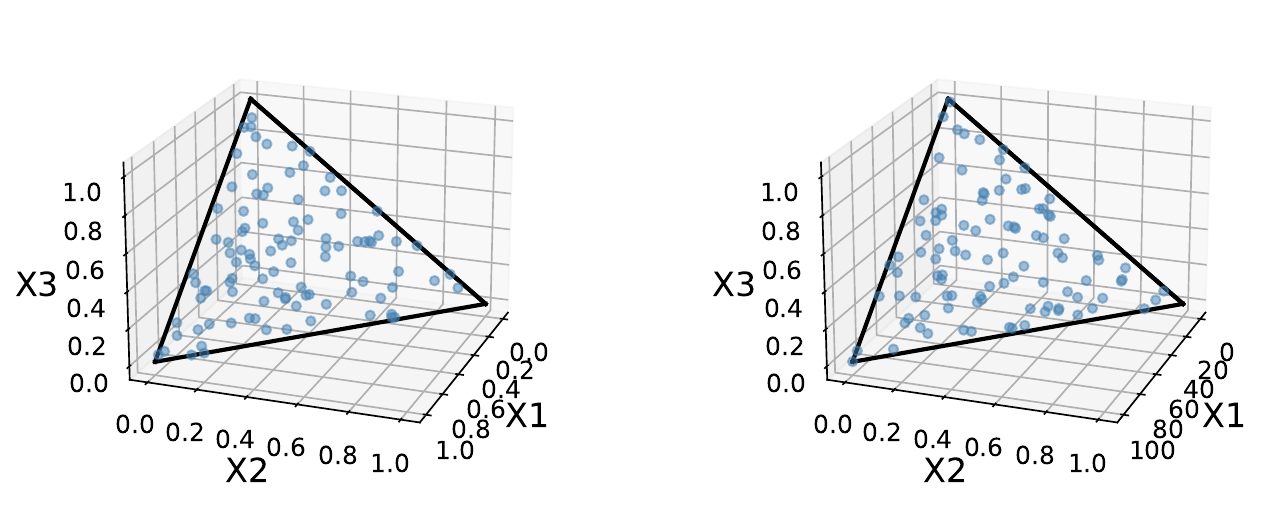}
    \caption{\textbf{Left}: $100$ samples drawn uniformly from within the unit simplex. \textbf{Right:} $100$ samples drawn from a non-unit simplex (note the scale of the $X1$ axis). The distribution of points in both simplexes is visually indistinguishable --- evidence that the method for sampling from a unit simplex is sufficient to draw samples from arbitrary simplexes.}
    \label{fig: sample-bias}
\end{figure}

Furthermore, Figure \ref{fig: sample-bias} shows that the distribution of samples in a unit simplex is visually similar to the samples from an elongated simplex where we multiply one of the basis vectors by a factor of $100$. This figure serves to show that although there may be some bias in our estimate of the loss over the simplex in Equation \ref{eqn:reg-loss}, it should not be (and is not in practice) limiting to our optimization routine. 
Note too, this may appear like a simplistic case, but typically the simplexes found by SPRO contain only a small number of vertices, so a $2$-simplex whose edge lengths vary by a factor of nearly $100$ is a reasonable comparison to a scenario we may find in practice. 

\paragraph{Computing Simplex Volume:} Simplex Volumes can be easily computed using Cayley-Menger determinants \citep{simplexVolume}. If we have an $n$-simplex defined by the parameter vectors $w_0, \dots, w_n$ the Cayley-Menger determinant is defined as 
\begin{equation}
    CM(w_0, \dots, w_n) =  
    \left| \begin{array}{ccccc}
    0 & d_{01}^{2} & \cdots & d_{0n}^{2} & 1\\
    d_{01}^{2} & 0 & \cdots & d_{0n}^{2} & 1\\
    \vdots & \vdots & \ddots & \vdots & \vdots \\
    d_{n0}^{2} & d_{n1}^{2} & \cdots & 0 & 1\\
    1 & 1 & 1 & 1 & 1\\
     \end{array} \right|.
\end{equation}

The Volume of the simplex $S_{(w_0, \dots, w_n)}$ is then given as
\begin{equation}
    \text{V}(S_{w_0, \dots, w_n)})^2 = \frac{(-1)^{n+1}}{(n!)^2 2^n}CM(w_0, \dots, w_n).
\end{equation}
While in general we may be adverse to computing determinants or factorial terms the simplexes we work with in this paper are generally low order (all are under $10$ vertices total) meaning that computing the Cayley-Menger determinants is generally a quite fast and stable computation.

\subsection{Initialization and Regularization}
\label{app: methods-details}
\paragraph{Vertex Initialization:} We initialize the $j^{th}$ parameter vector corresponding to a vertex in the simplex as the mean of the previously found vertices, $w_j = \frac{1}{j}\sum_{i=0}^{j-1}w_i$ and train using the regularized loss in Eq. \ref{eqn:reg-loss}.

\paragraph{Regularization Parameter:}
As the order of the simplex increases, the Volume of the simplex increases exponentially.
Thus, we define a distinct regularization parameter, $\lambda_j$, in training each $\theta_j$ to provide consistent regularization for all vertices.
To choose the $\lambda_k$'s we define a $\lambda^*$ and compute
\begin{equation}\label{eqn:reg-par}
	\lambda_k = \frac{\lambda^*}{\log \text{V}(\mathcal{K})},
\end{equation}
where $\mathcal{K}$ is randomly initialized simplicial complex of the same structure that the simplicial complex will have while training $\theta_j$. Eq. \ref{eqn:reg-par} normalizes the $\lambda_k$'s such that they are similar when accounting for the exponential growth in volume as the order of the simplex grows.
In practice we need only small amounts of regularization, and choose $\lambda^*=10^{-8}$. As we are spanning a space of near constant loss any level of regularization will encourage finding simplexes with non-trivial Volume.

\begin{figure}
    \centering
    \includegraphics[width=0.6\linewidth]{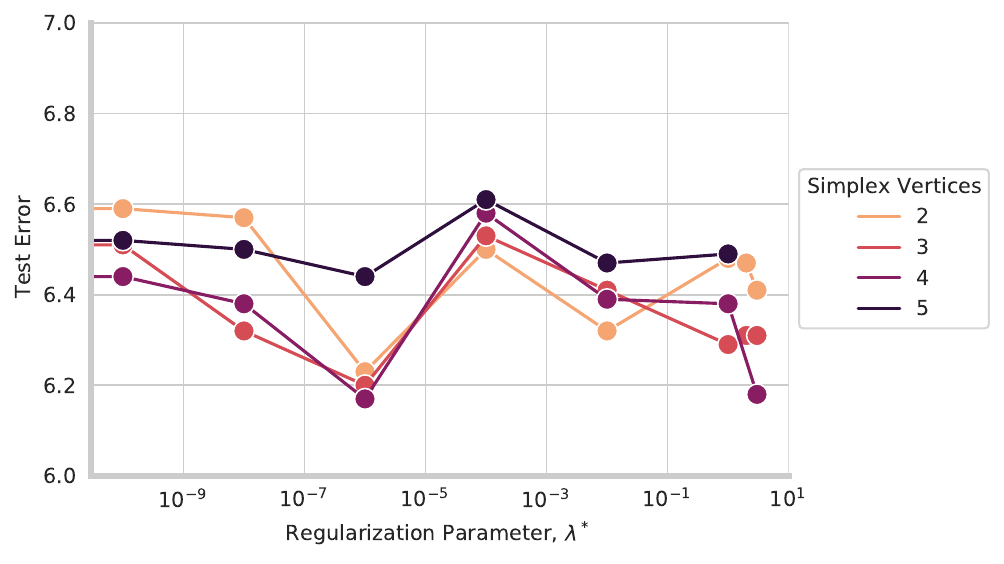}
    \caption{CIFAR-$10$ test accuracy as a function of regularization parameter $\lambda^*$ and colored by the number of vertices. Accuracy is essentially unchanged for the various regularization parameters.}
    \label{fig:regularization_ablation}
\end{figure}

Finally, when dealing with models that use batch normalization, we follow the procedure of \citet{garipov2018loss} and compute several forwards passes on the training data for a given sample from the simplex to update the batch normalization statistics. 
For layer normalization, we do not need to use this procedure as layer norm updates at test time.

\subsection{Training Details}
We used VGG-16 like networks originally introduced in \citet{simonyan2014very} from \url{https://github.com/pytorch/vision/blob/master/torchvision/models/vgg.py}.
For training, we used standard normalization, random horizontal flips, and crops and a batch size of $128$.
We used SGD with momentum $=0.9,$ and a cosine annealing learning rate with a single cycle, a learning rate of $0.05,$ and weight decay $5e-4$ training for $300$ epochs for the pre-trained VGG models.
For SPRO, we used a learning rate of $0.01$ and trained for $20$ epochs for each connector.

In our experiments with transformers, we used the \emph{ViT-B\_16} image transformer model \citep{dosovitskiy2020image} pre-trained on ImageNet from \url{https://github.com/jeonsworld/ViT-pytorch} and trained on CIFAR100 with upsampled image size of $224$ with a batch size of $512$ for $50000$ steps (the default fine-tuning on CIFAR-100). 
Again, we used random flips and crops for data augmentation.
To train these SPRO models, we used a learning rate of $0.001$ and trained with SGD for $30$ epochs for each connector, using $20$ samples from the simplex at test time.

\subsection{Multi-Dimensional Mode Connectors}\label{app:swag_connectors}
To train the multi-dimensional SWAG connectors, we connected two pre-trained networks following \citet{garipov2018loss} using a piece-wise linear curve, trained for $75$ epochs with an initial learning rate of $0.01,$ decaying the learning rate to $1e-4$ by epoch $40.$
At epoch $40,$ we reset the learning rate to be constant at $5e-3.$
The final individual sample accuracy (not SWA) was $91.76\%,$ which is similar to the final individual sample accuracies for standard training of VGG networks with SWAG.
We used random crops and flips for data augmentation.

\section{Extended Volume and Ensembling Results}\label{app:espro}
\subsection{Volumes on MNIST}\label{app:mnist_volumes}

In a similar construction to the dimensionality experiment in Figure \ref{fig: conn-limit-vol}, we next consider lower bounding the dimensionality of the connecting space that SPRO can find for LeNet-$5$s on MNIST \citep{lecun1998gradient}\footnote{Implementation from \url{https://github.com/activatedgeek/LeNet-5/blob/master/lenet.py}}, varying the width of the convolutional networks from a baseline of $1$ (standard parameterization), either halving the width or consecutively widening the layers by a constant factor.
We find in Figure \ref{fig:mnist_vols} that the volumes of the simplicial complex can vary by several powers of $10$ for the as we increase the widths. 
However, all width networks generally follow the same patter of decaying volume as we increase the number of connecting points (e.g. increasing the dimensionality of the simplicial complex).

\subsection{Test Error vs. Simplex Samples}\label{app:error-v-sample}

SPRO gives us access to a whole space of model parameters to sample from rather than just a discrete number of models to use as in deep ensembles. Therefore a natural question to ask is how many models and forwards passes need to be sampled from the simplex to achieve the highest accuracy possible without incurring too high of a cost at test time. 

\begin{figure}
\centering
    \begin{subfigure}{0.49\textwidth}
    \centering
        \includegraphics[width=0.8\linewidth]{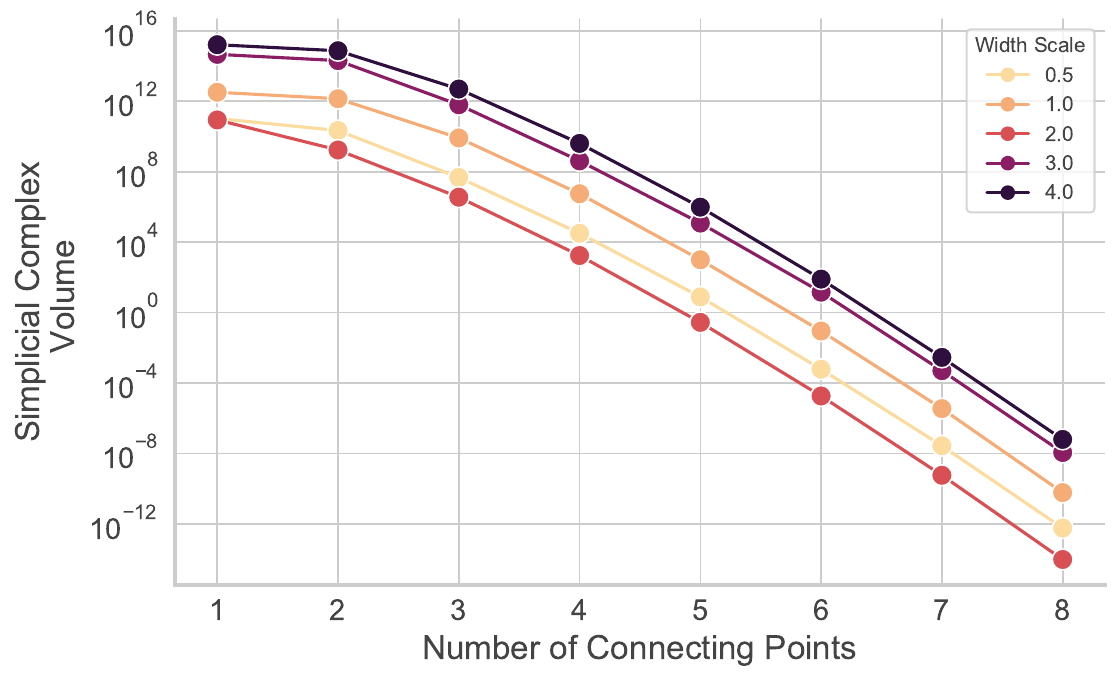}
        \caption{Log volumes, MNIST.}
        \label{fig:mnist_vols}
    \end{subfigure}
    \begin{subfigure}{0.49\textwidth}
    \centering
        \includegraphics[width=0.8\linewidth]{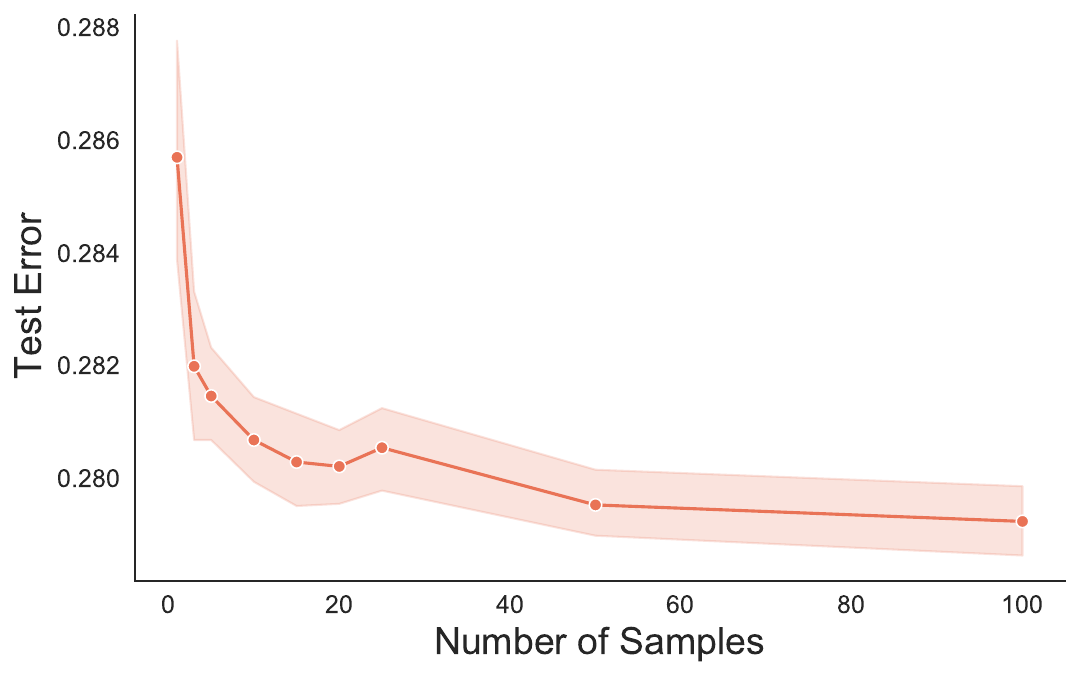}
        \caption{Test error vs. number of samples, CIFAR-$100$}
        \label{fig:error-v-sample}
    \end{subfigure}

    \caption{\textbf{(a)} Log volumes as a function of LeNet-$5$ layer width. Volumes are generally highest for wider models, and the volume of the simplicial complex tends to decrease as the dimension of the space increases. \textbf{(b)}Test error vs. number of samples, $J,$ in the ensemble on CIFAR-$100$ using a VGG-$16$ network and a $3$-simplex trained with SPRO. For any number of components in the SPRO ensemble greater than approximately $25$ we achieve near constant test error.}
    
\end{figure}

Figure \ref{fig:error-v-sample} shows that for a VGG-$16$ network trained on CIFAR-$100$ we achieve near constant accuracy for any number of ensemble components greater than approximately $25$. Therefore, for the ensembling experiments in Section \ref{sec: ensembles} we use $25$ samples from each simplex to generate the SPRO ensembles. In this work we are not focused on the issue of test time compute cost, and if that were a consideration for deployment of a SPRO model we could evaluate the trade-off in terms of test time compute vs accuracy, or employ more sophisticated methods such as ensemble distillation.

\subsection{Loss Surfaces of Transformers}\label{app:transformer_ls}
\begin{figure}[t!]
    \centering
    \includegraphics[width=\linewidth]{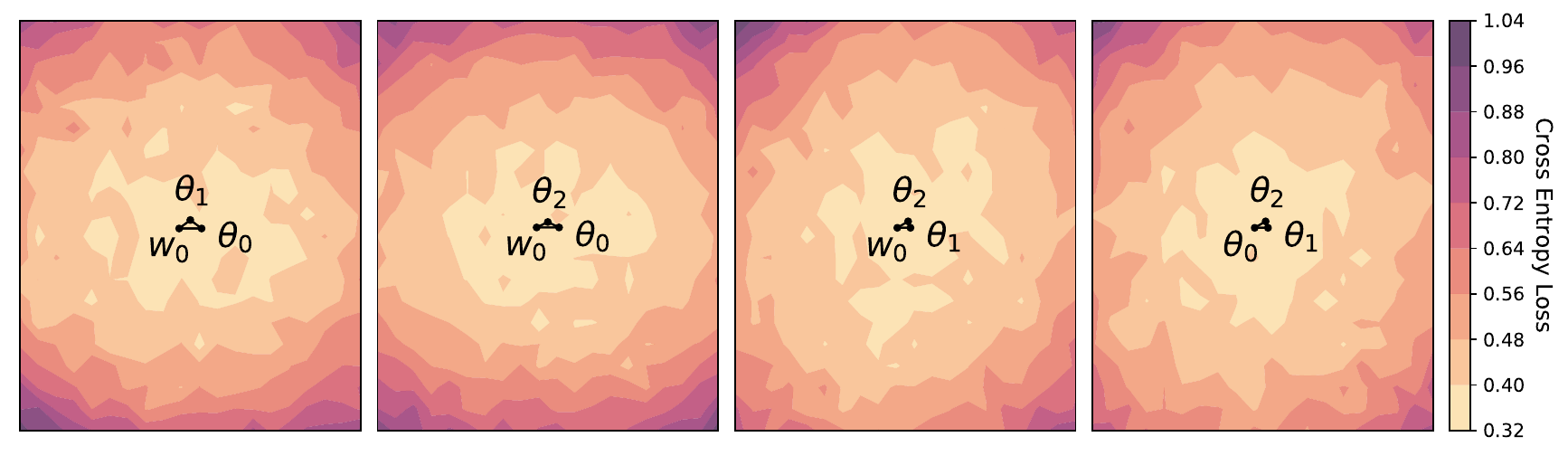}
    \caption{Loss surface visualizations of the faces of a sample ESPRO $3$-simplex for a Transformer architecture \citep{dosovitskiy2020image} fine-tuned on CIFAR-$100$. Here, the volume is considerably smaller, but a low loss region is found.}
    \label{fig:c100-espro-simplex-transformer}
\end{figure}

Next, we show the results of training a SPRO $3-$simplex with an image transformer on CIFAR-$100$ \citep{dosovitskiy2020image} in Figure \ref{fig:c100-espro-simplex-transformer}.
Due to computational requirements, the transformer was pre-trained on ImageNet before being fine-tuned on CIFAR-$100$ for $50,000$ minibatches.
We then trained each vertex for an additional $10$ epochs.
Due to the inflexibility of the architecture, we observed training instability, which ultimately produced a small volume of the simplex found (approximately $10^{-21}$).
Furthermore, the small volume of the simplex produced less diverse solutions, limiting the benefits of ensembling transformer models
as shown in Figure \ref{fig: transformers-deep-ensembles}.
However, these results demonstrate that a region of low loss can be found in subspaces of transformer models, and further work will be necessary to efficiently exploit these regions of low loss, much like has been done with CNNs and ResNets.

\subsection{Ensembling Mode Connecting Simplexes}\label{app:mode_espro}
\begin{figure}
    \centering
    \includegraphics[width=\linewidth]{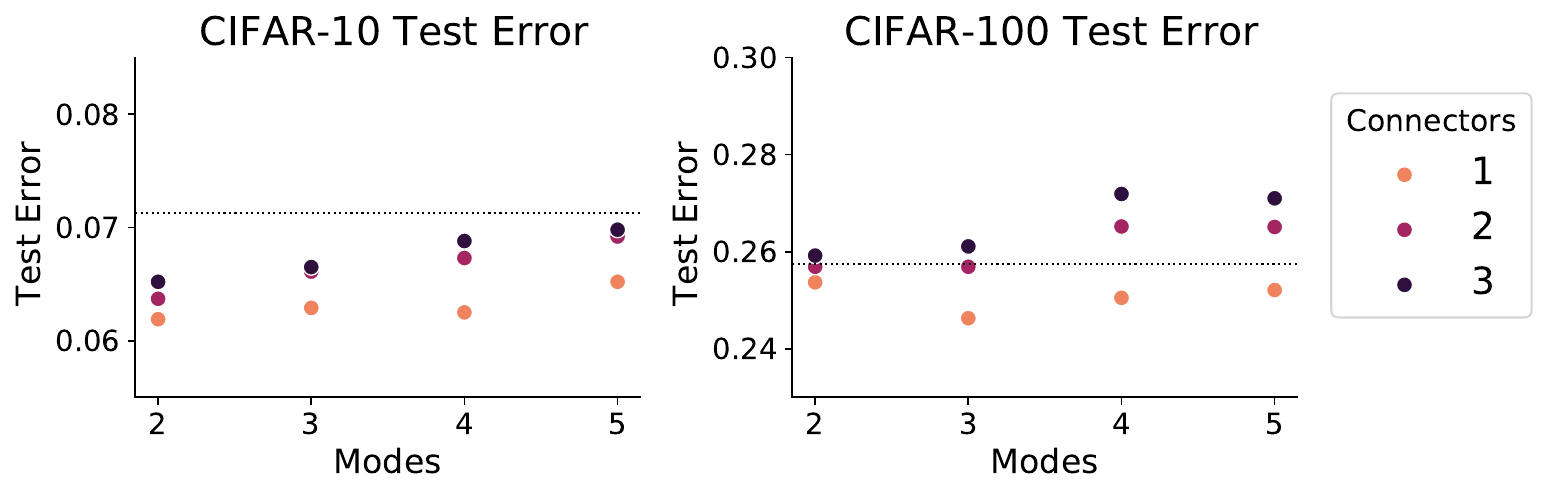}
    \caption{Test error for mode connecting simplexes that connect various numbers of modes through various numbers of connecting points in the parameter space of VGG-$16$ networks trained on CIFAR-$10$ and CIFAR-$100$. The error rates of baseline models are shown as horizontal dotted lines. In general the highest performing models are those with the fewest modes and the fewest connecting points, but the performance gaps between configurations are small.}
    \label{fig:connectivity-accs}
\end{figure}

We can average predictions over these mode connecting volumes, generating predictions as ensembles, $\hat{y} = \frac{1}{H}\sum_{\phi_h \sim \mathcal{K}}f(x, \phi_h)$, where $\phi_h \sim \mathcal{K}$ indicates we are sampling models uniformly at random from the simplicial complex $\mathcal{K}(S_{(w_0, \theta_0, \dots)}, S_{(w_1, \theta_0, \dots)}, \dots)$. Test error for such ensembles for volumes in the parameter space of VGG-$16$ style networks on both CIFAR-$10$ and CIFAR-$100$ are given in Figure \ref{fig:connectivity-accs}.
We see that while some improvements over baselines can be made, mode connecting simplexes do not lead to highly performant ensembles.

\subsection{Ensembling Modes of SPRO}
\begin{figure*}
    \centering
    \includegraphics[width=\linewidth]{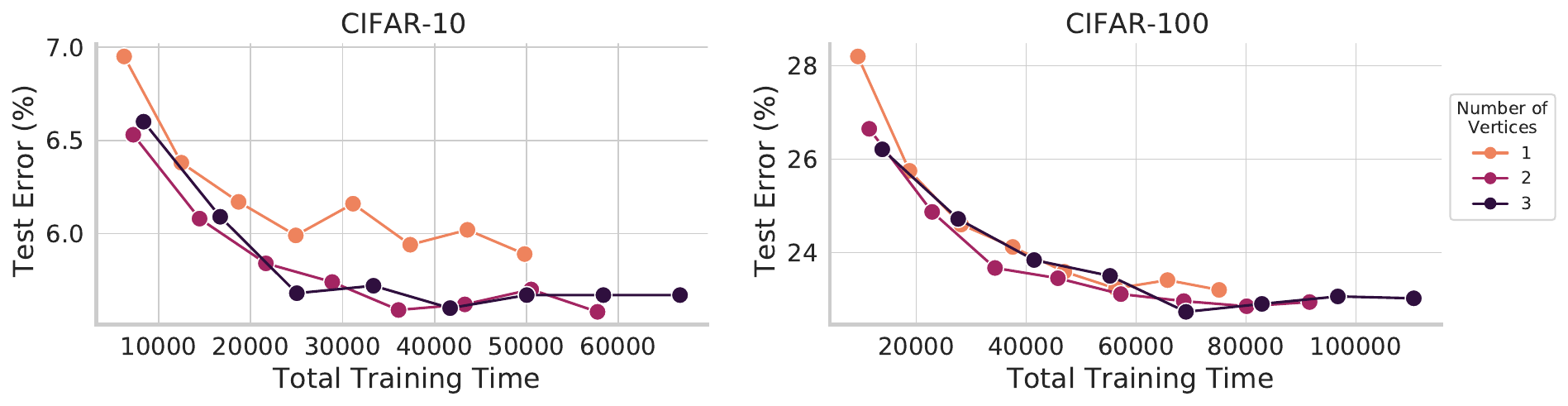}
    \caption{Test error of ESPRO models on CIFAR-$10$ (left) and CIFAR-$100$ (right) as a function of total training time (training the original models and the ESPRO simplexes). The color of the curves indicate the number of the vertices in the simplex, and the points corresponding to increasing numbers of ensemble components moving left to right (ranging from $1$ to $8$). We see that on either dataset for nearly any fixed training budget, we are better off training fewer models overall and using ESPRO to construct simplexes to sample from.}
    \label{fig:espro-acc-time}
\end{figure*}

Figure \ref{fig:espro-acc-time} presents the results of Figure \ref{fig: deep-ensembles} in the main text, but against the total training budget rather than the number of ensemble components. We see from the plot that on either dataset, for nearly any fixed training budget the most accurate model is an ESPRO model, even if that means using our budget to train ESPRO simplexes but fewer models overall.

Times correspond to training models sequentially on an NVIDIA Titan RTX with $24GB$ of memory.

\begin{figure*}[t!]
	\centering
	\includegraphics[width=\linewidth]{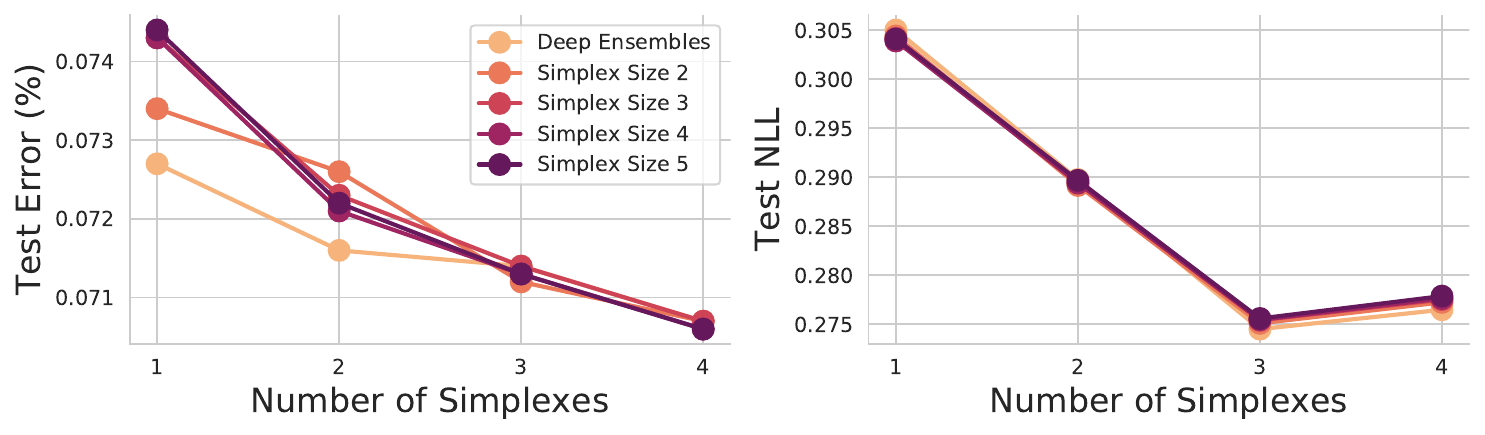}
	\caption{Test Error and NLL for the number of components in SPRO ensembles using image transformers on CIFAR-100. ESPRO with four dimensional simplexes is slightly better in test accuracy and slightly worse in test NLL than deep ensembles.}
	\label{fig: transformers-deep-ensembles}
\end{figure*}

Finally, Figure \ref{fig: transformers-deep-ensembles} presents the results of ensembling with SPRO using state of the art transformers architectures on CIFAR-100 \citep{dosovitskiy2020image}. 
We find, counterintuitively that there is only a very small performance difference from ensembling with SPRO compared to the base architecture.
We suspect that this is because it is currently quite difficult to train transformers without using significant amounts of unlabelled data.

\section{Extended Uncertainty Results}\label{app:reg_uncertainty}

\subsection{Further NLL and Calibration  Results}\label{app:nll_calib}

Finally, we include the results across $18$ different corruptions for the ensemble components.
In order, these are \textit{jpeg}, fog, snow, brightness, pixelate, zoom blur, saturate, contrast, motion blur, defocus blur, speckle noise, gaussian blur, glass blur, shot noise, frost, spatter, impulse noise and, elastic transform. 

\begin{figure*}
    \centering
    \includegraphics[width=\linewidth]{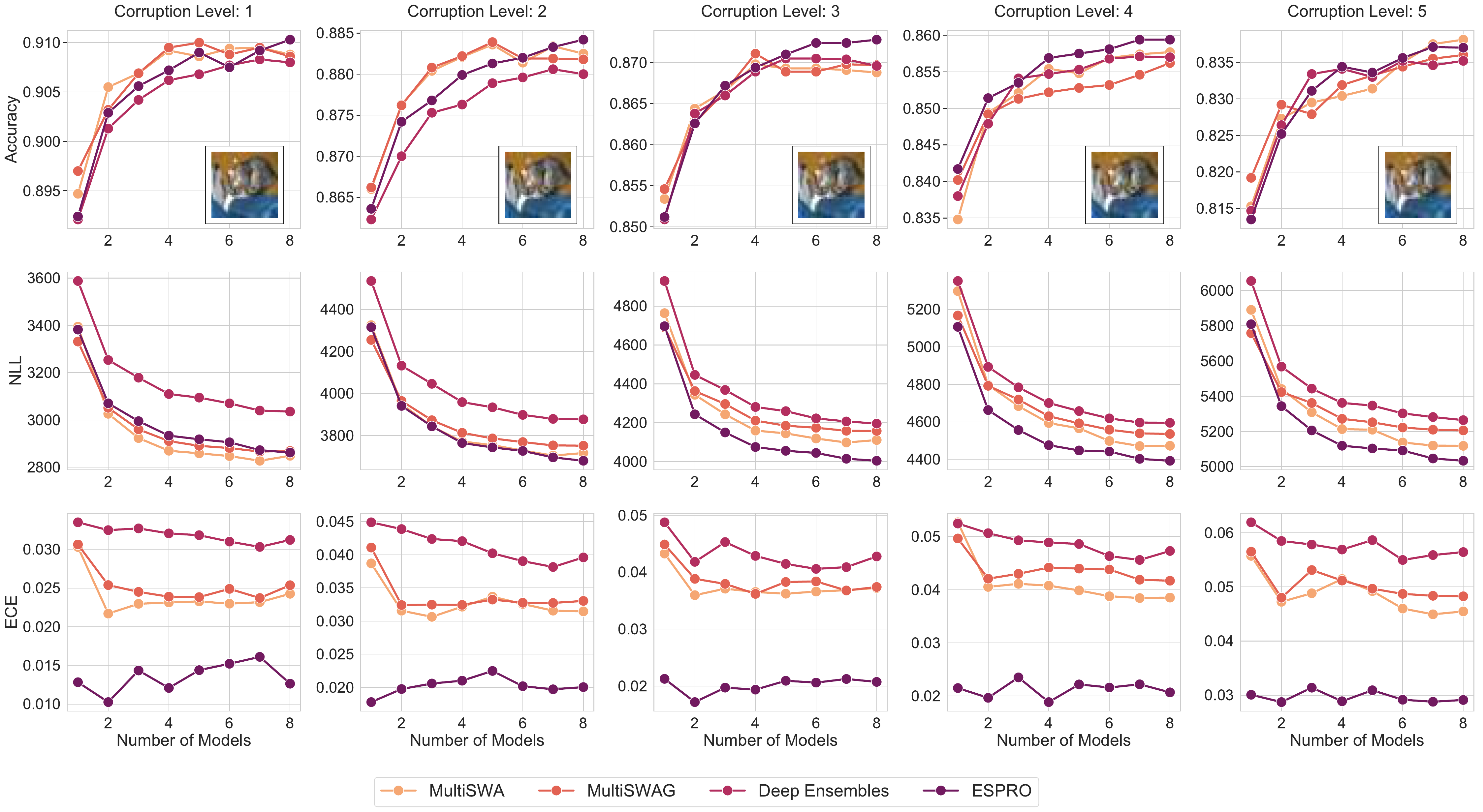}
    \caption{Accuracy, NLL and ECE with increasing intensity of the \textit{jpeg compression} corruption (from left to right).}
    \label{fig:jpeg_corruption}
\end{figure*}

\begin{figure*}
    \centering
    \includegraphics[width=\linewidth]{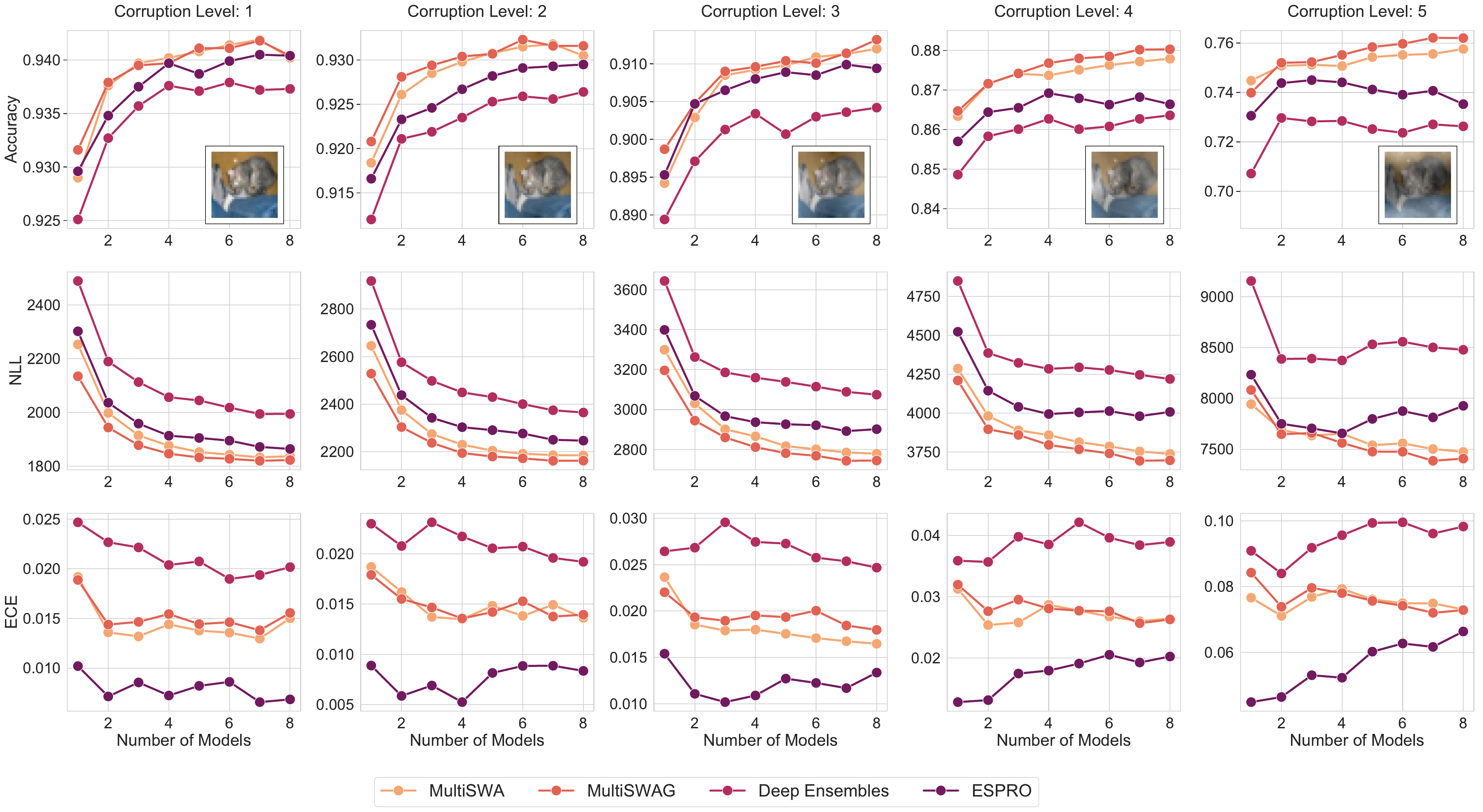}
    \caption{Accuracy, NLL and ECE with increasing intensity of the \textit{fog} corruption (from left to right).}
    \label{fig:fog_corruption}
\end{figure*}

\begin{figure*}
    \centering
    \includegraphics[width=\linewidth]{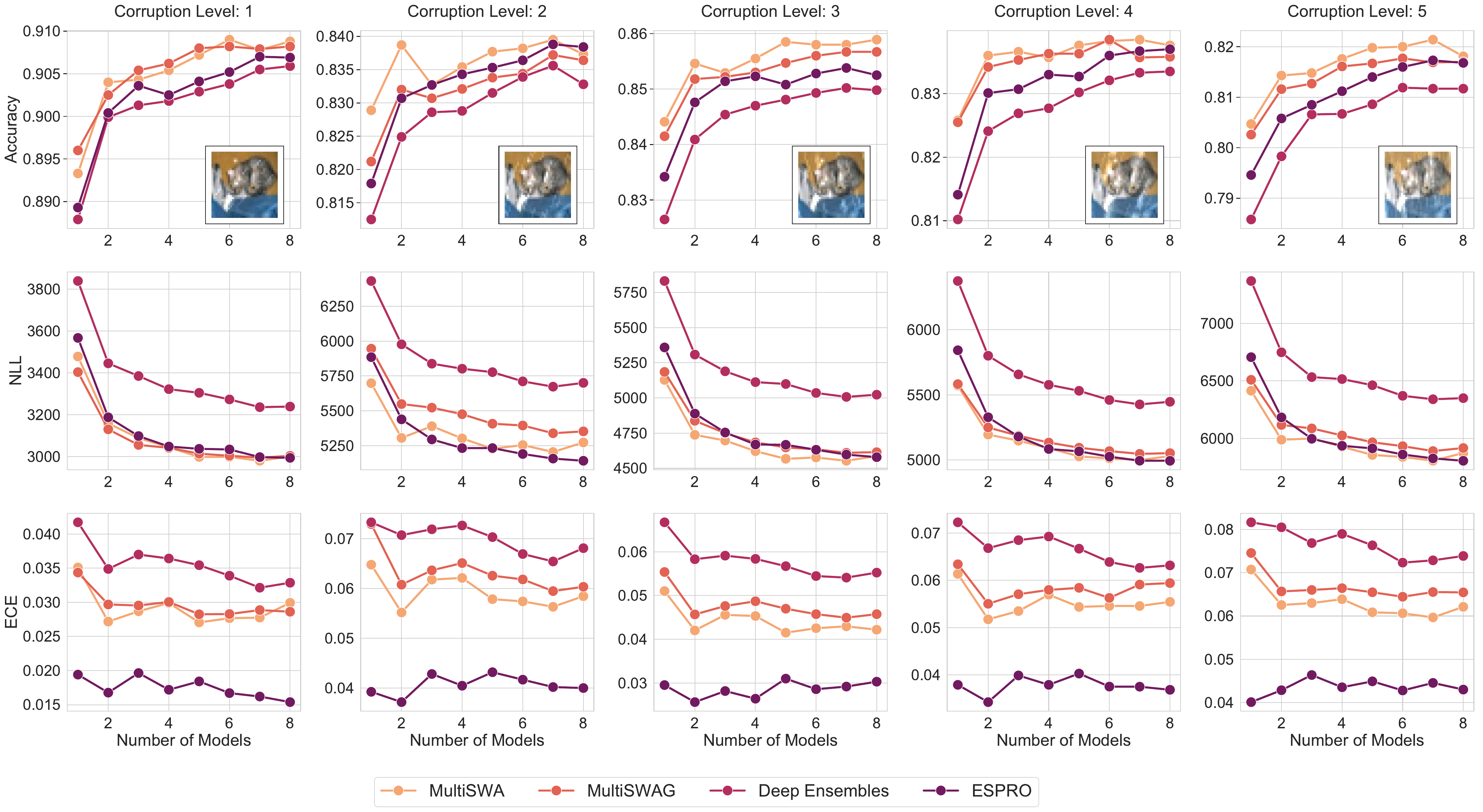}
    \caption{Accuracy, NLL and ECE with increasing intensity of the \textit{snow} corruption (from left to right).}
    \label{fig:snow_corruption}
\end{figure*}

\begin{figure*}
    \centering
    \includegraphics[width=\linewidth]{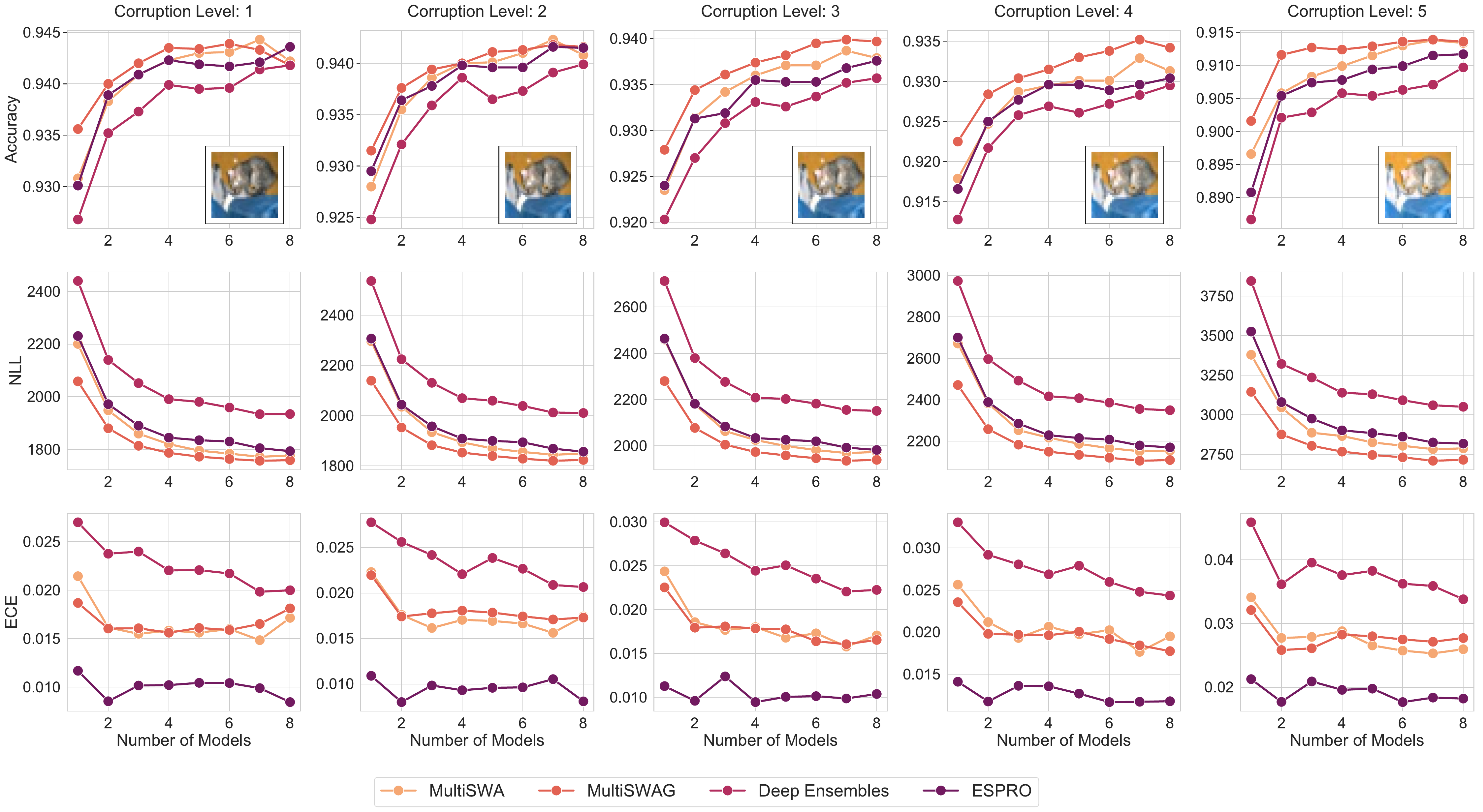}
    \caption{Accuracy, NLL and ECE with increasing intensity of the \textit{brightness} corruption (from left to right).}
    \label{fig:brightness_corruption}
\end{figure*}

\begin{figure*}
    \centering
    \includegraphics[width=\linewidth]{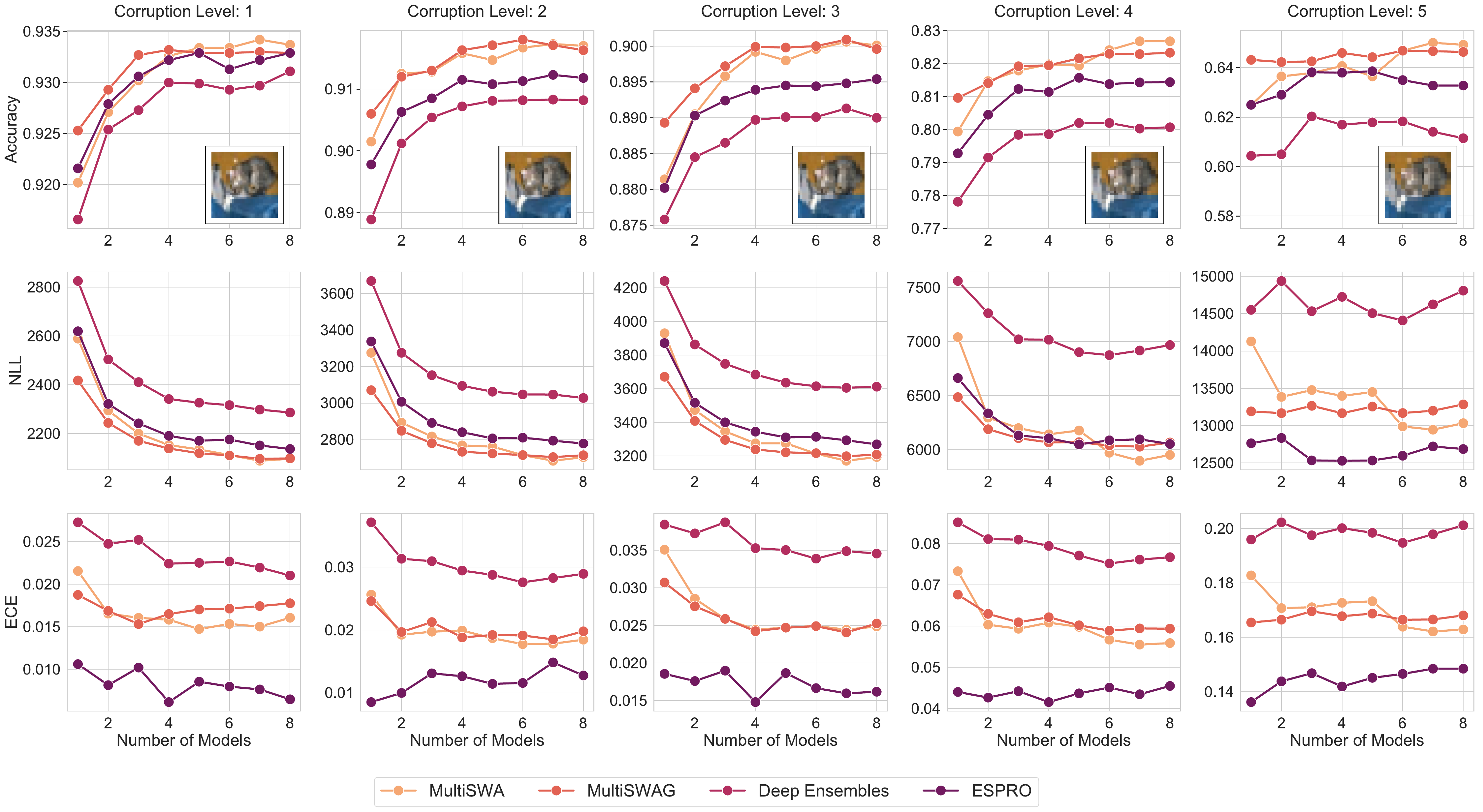}
    \caption{Accuracy, NLL and ECE with increasing intensity of the \textit{pixelate} corruption (from left to right).}
    \label{fig:pixelate_corruption}
\end{figure*}

\begin{figure*}
    \centering
    \includegraphics[width=\linewidth]{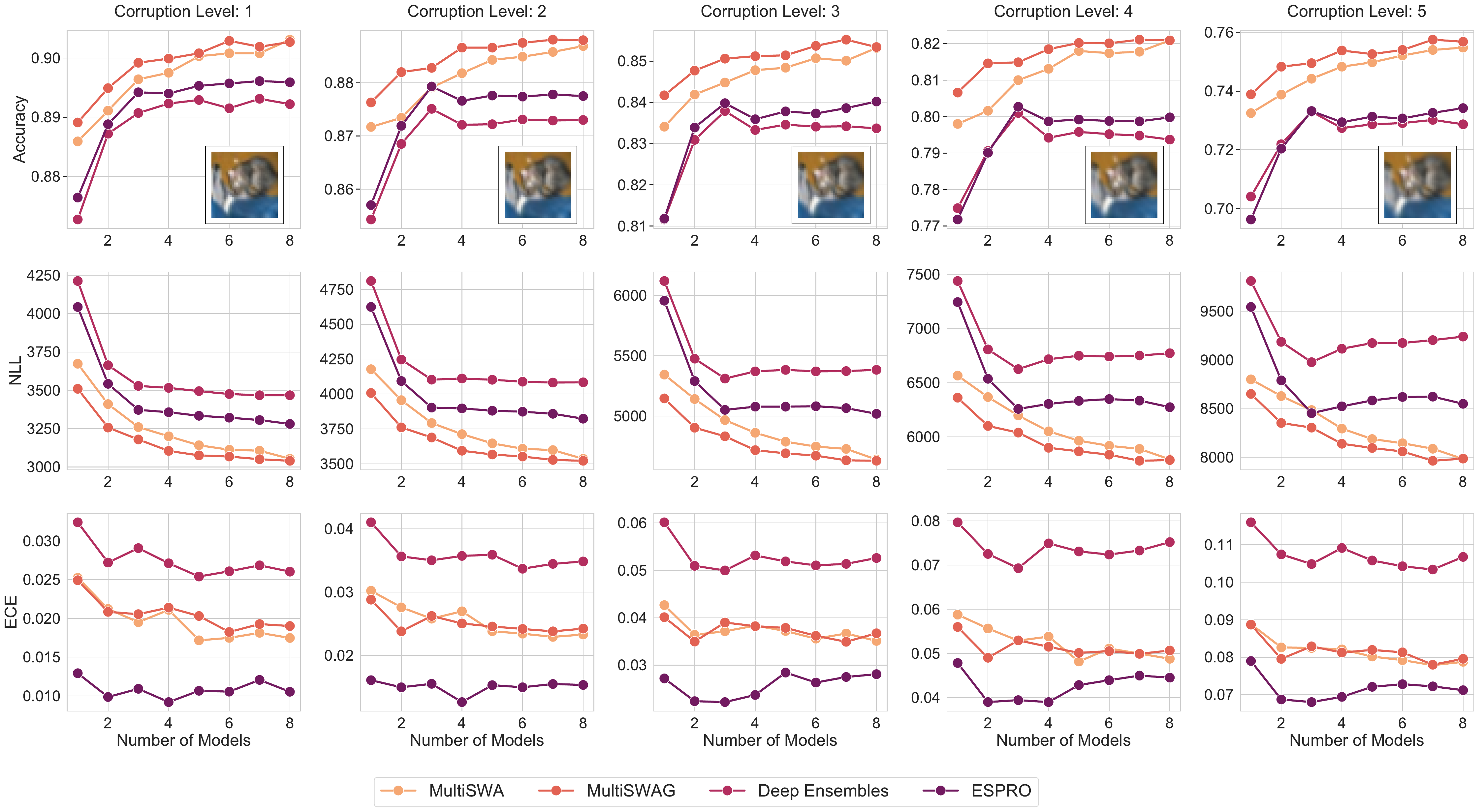}
    \caption{Accuracy, NLL and ECE with increasing intensity of the \textit{zoom blur} corruption (from left to right).}
    \label{fig:pixelate_corruption}
\end{figure*}

\begin{figure*}
    \centering
    \includegraphics[width=\linewidth]{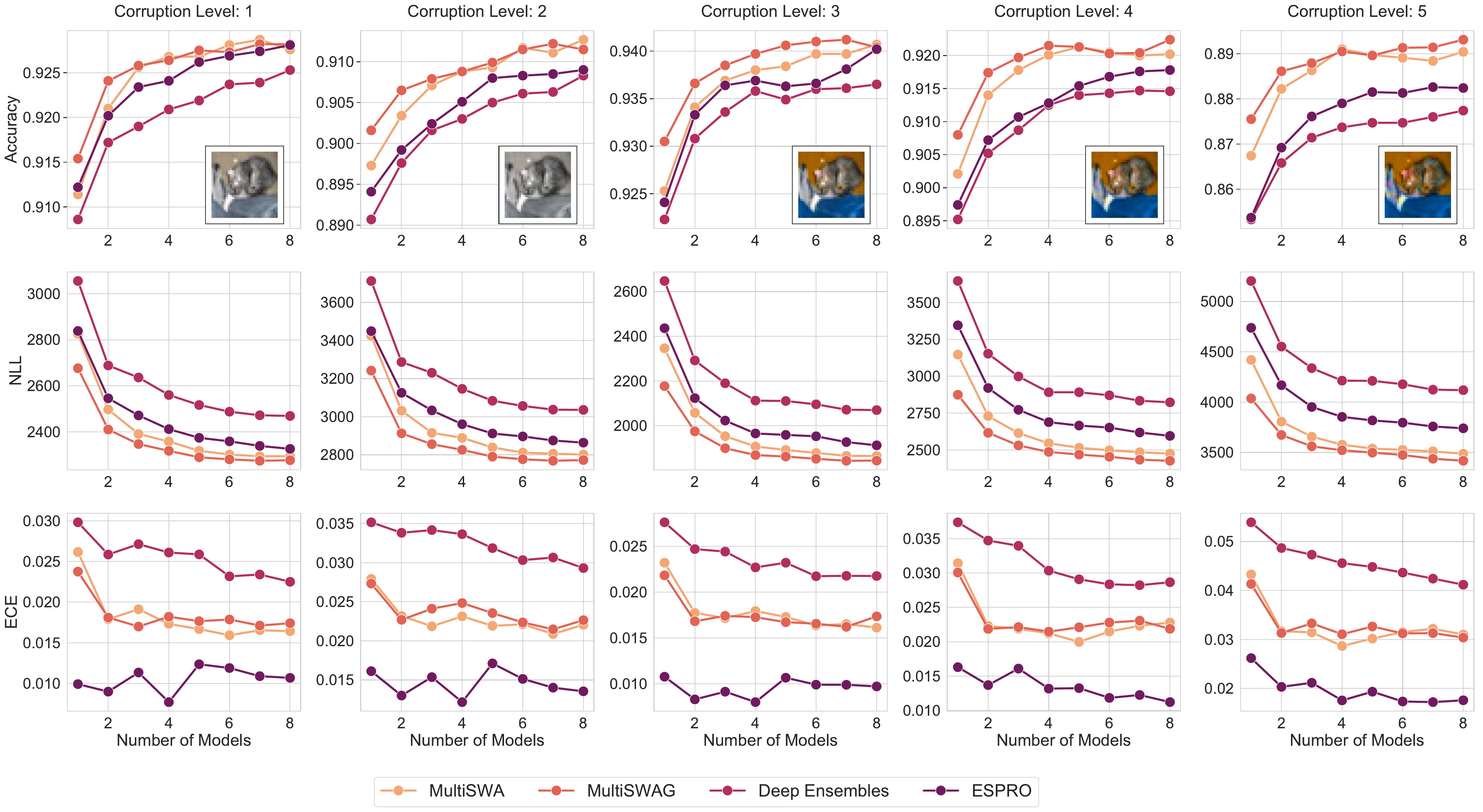}
    \caption{Accuracy, NLL and ECE with increasing intensity of the \textit{saturate} corruption (from left to right).}
    \label{fig:pixelate_corruption}
\end{figure*}

\begin{figure*}
    \centering
    \includegraphics[width=\linewidth]{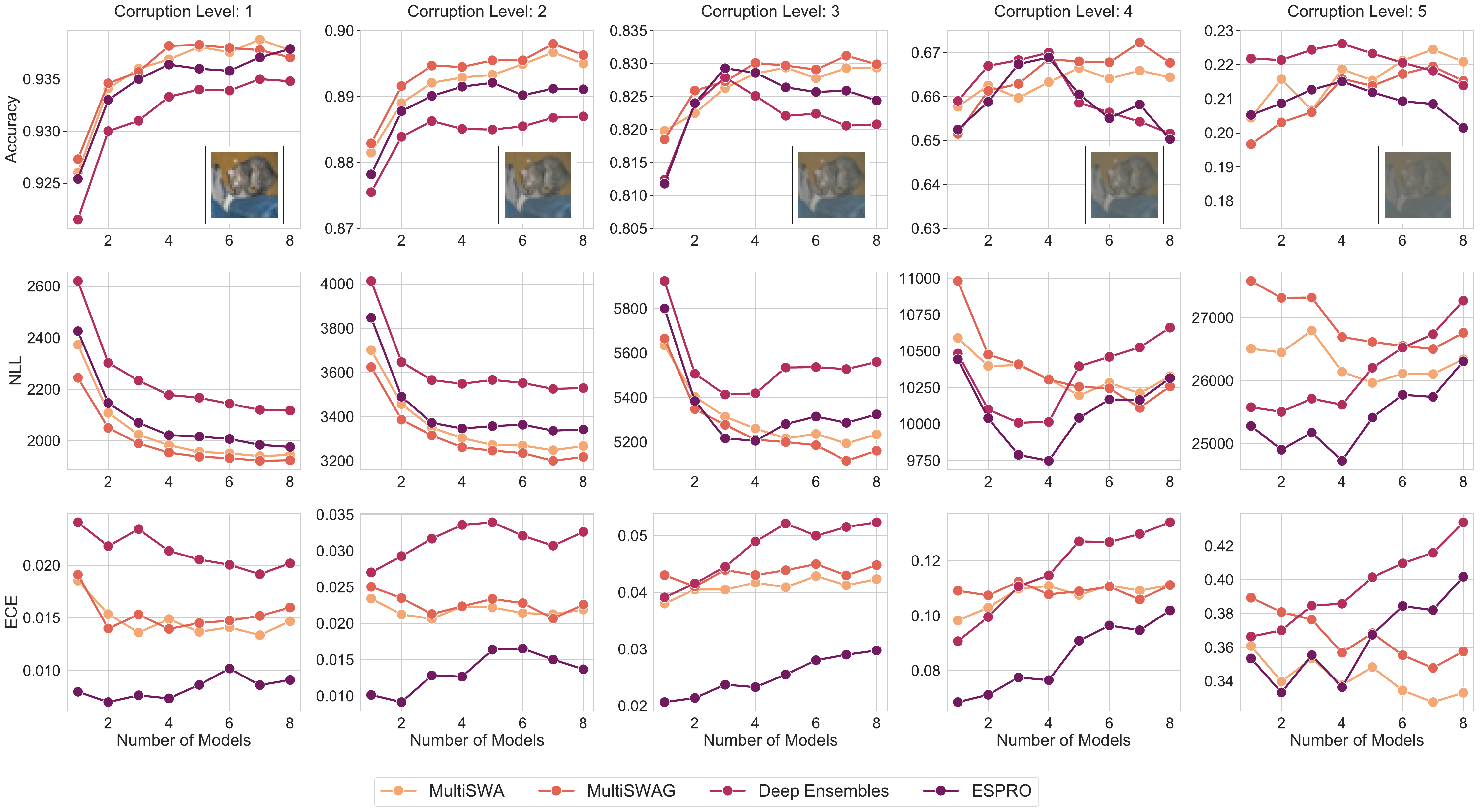}
    \caption{Accuracy, NLL and ECE with increasing intensity of the \textit{contrast} corruption (from left to right).}
    \label{fig:pixelate_corruption}
\end{figure*}

\begin{figure*}
    \centering
    \includegraphics[width=\linewidth]{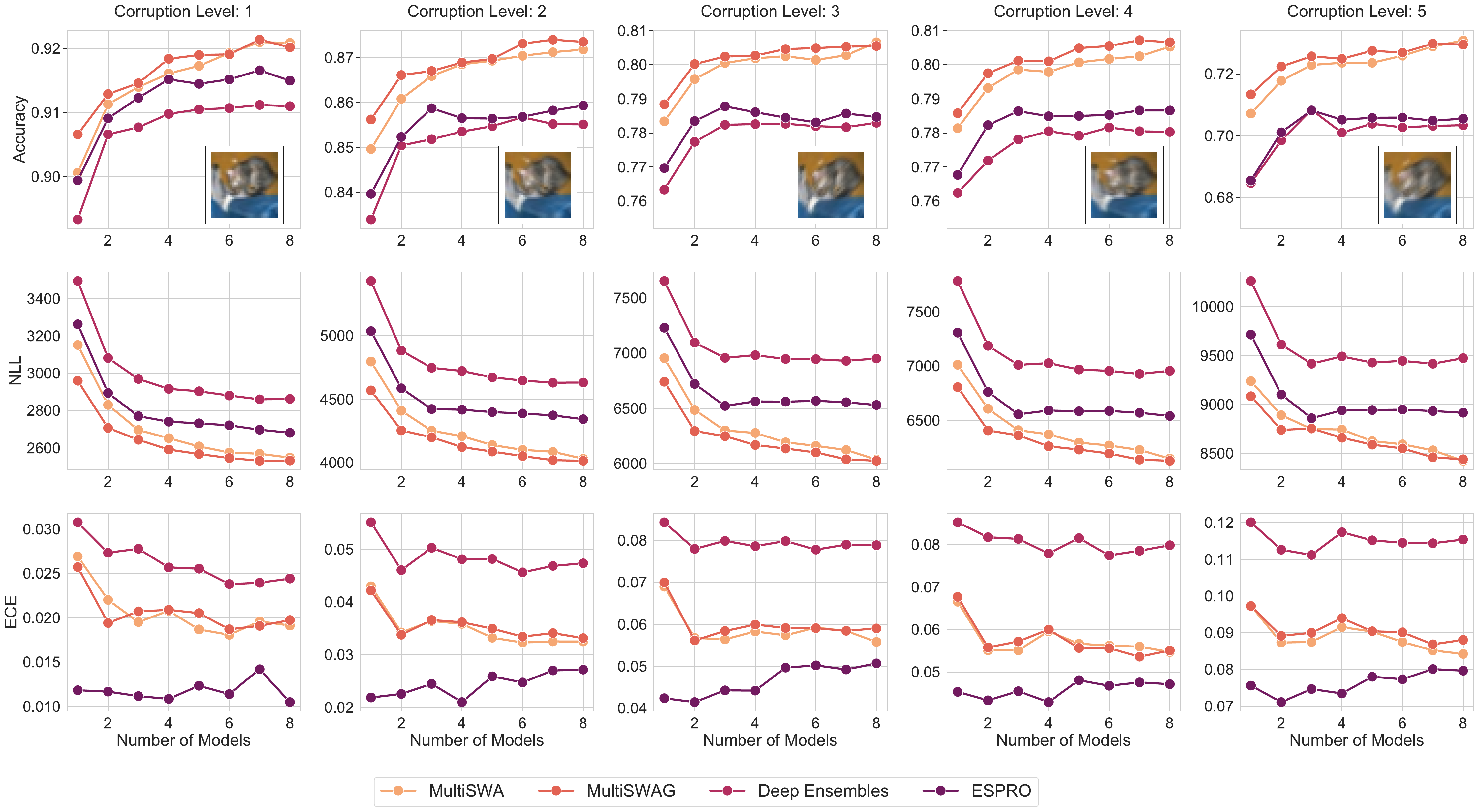}
    \caption{Accuracy, NLL and ECE with increasing intensity of the \textit{motion blur} corruption (from left to right).}
    \label{fig:pixelate_corruption}
\end{figure*}

\begin{figure*}
    \centering
    \includegraphics[width=\linewidth]{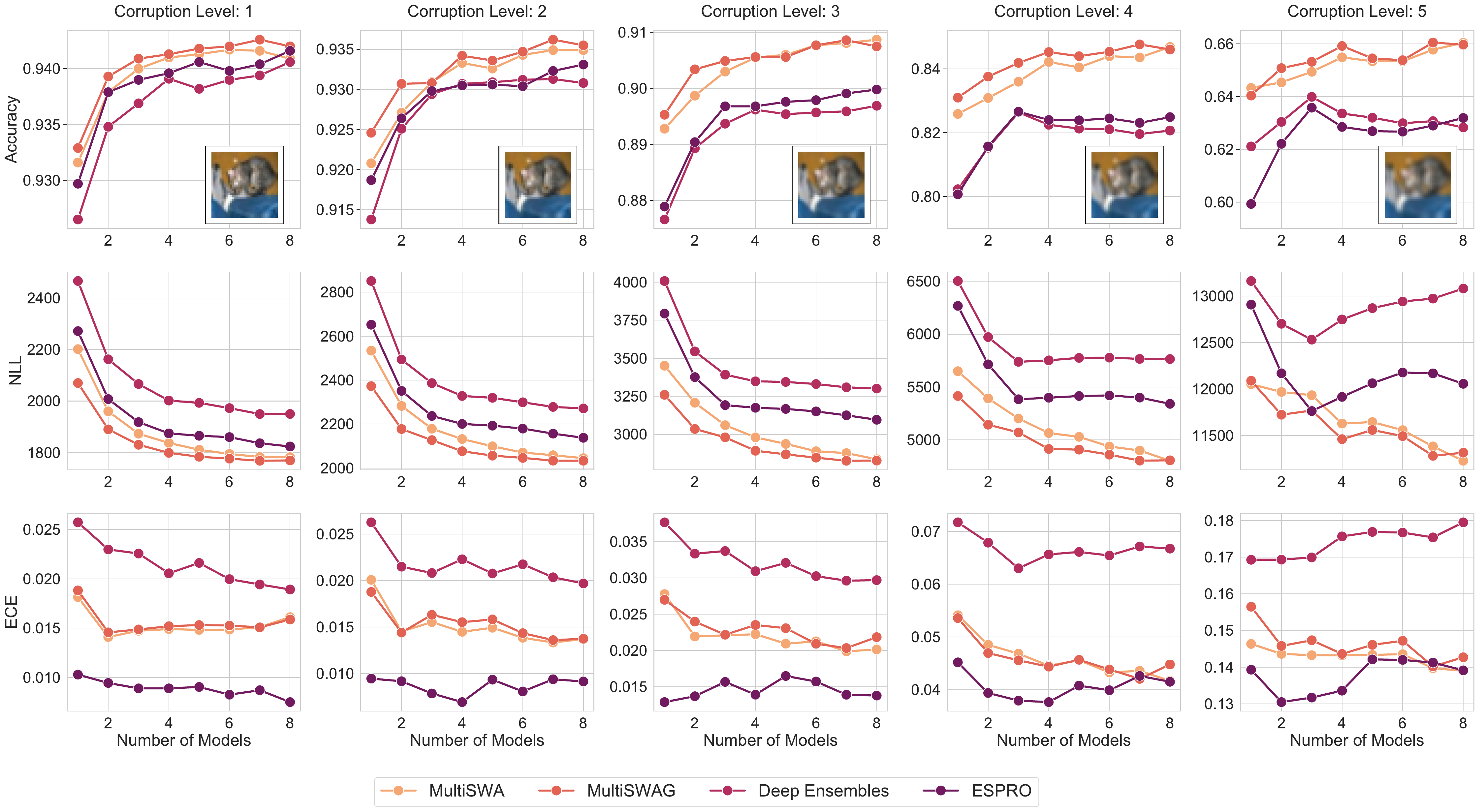}
    \caption{Accuracy, NLL and ECE with increasing intensity of the \textit{defocus blur} corruption (from left to right).}
    \label{fig:pixelate_corruption}
\end{figure*}

\begin{figure*}
    \centering
    \includegraphics[width=\linewidth]{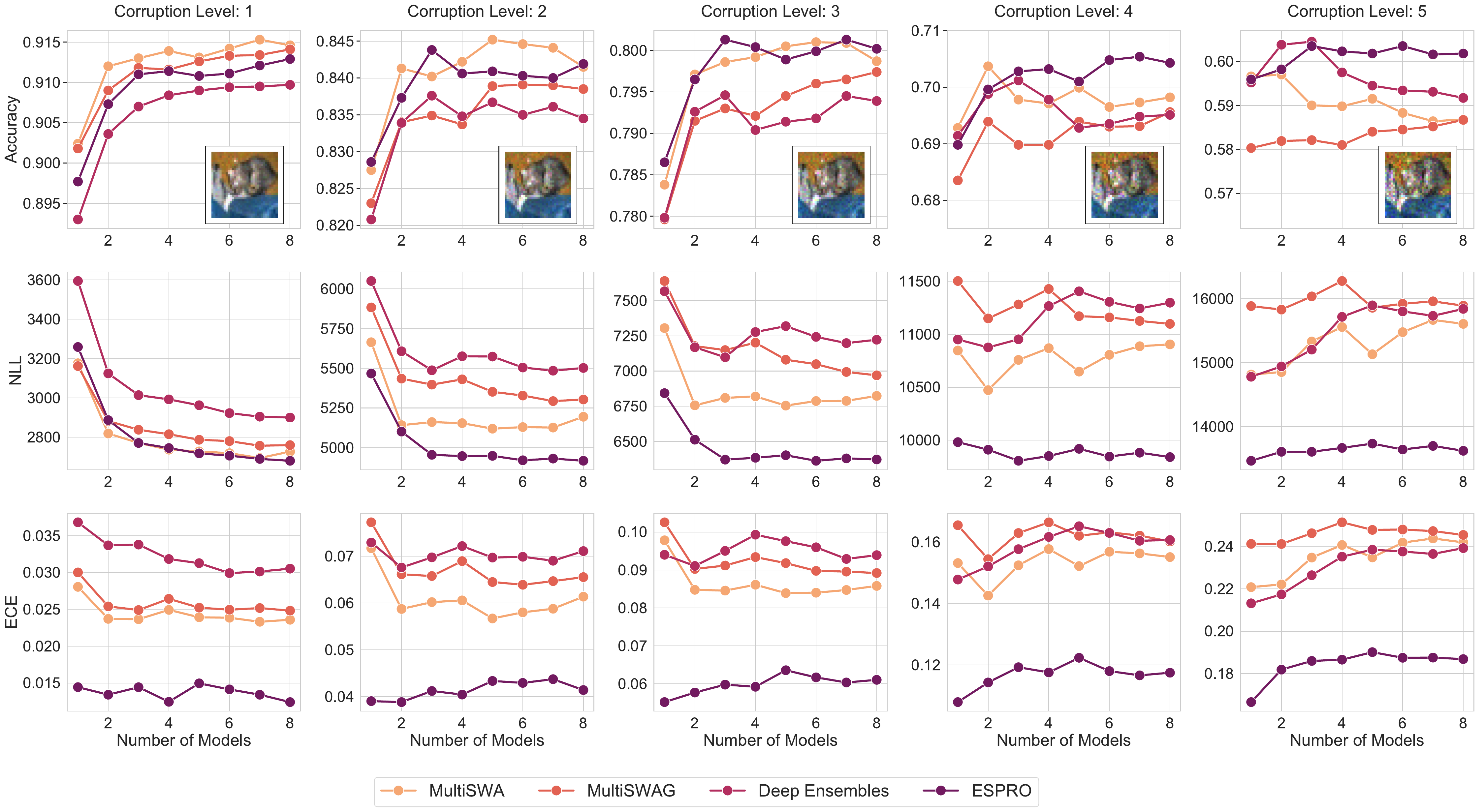}
    \caption{Accuracy, NLL and ECE with increasing intensity of the \textit{speckle noise} corruption (from left to right).}
    \label{fig:pixelate_corruption}
\end{figure*}

\begin{figure*}
    \centering
    \includegraphics[width=\linewidth]{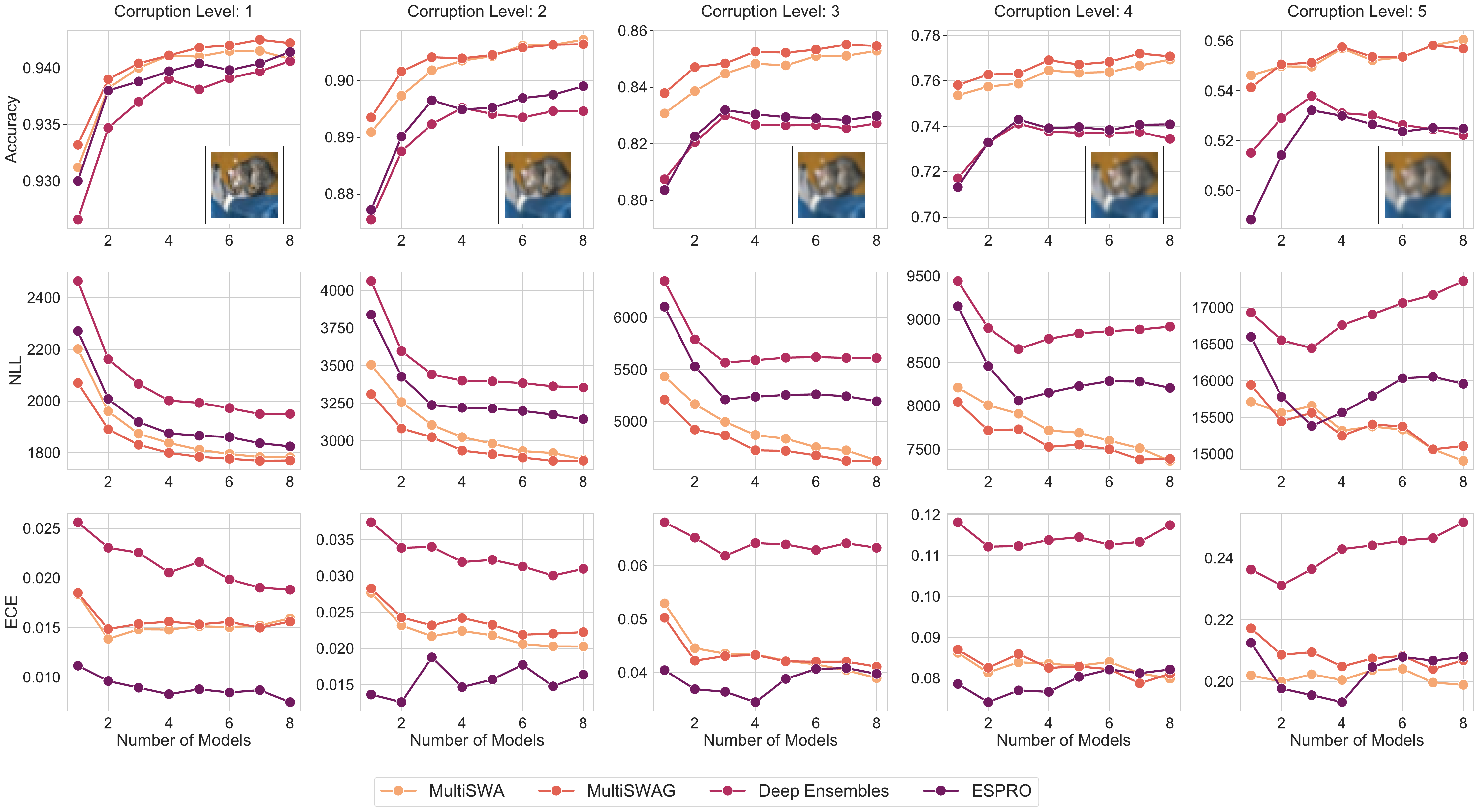}
    \caption{Accuracy, NLL and ECE with increasing intensity of the \textit{Gaussian blur} corruption (from left to right).}
    \label{fig:pixelate_corruption}
\end{figure*}

\begin{figure*}
    \centering
    \includegraphics[width=\linewidth]{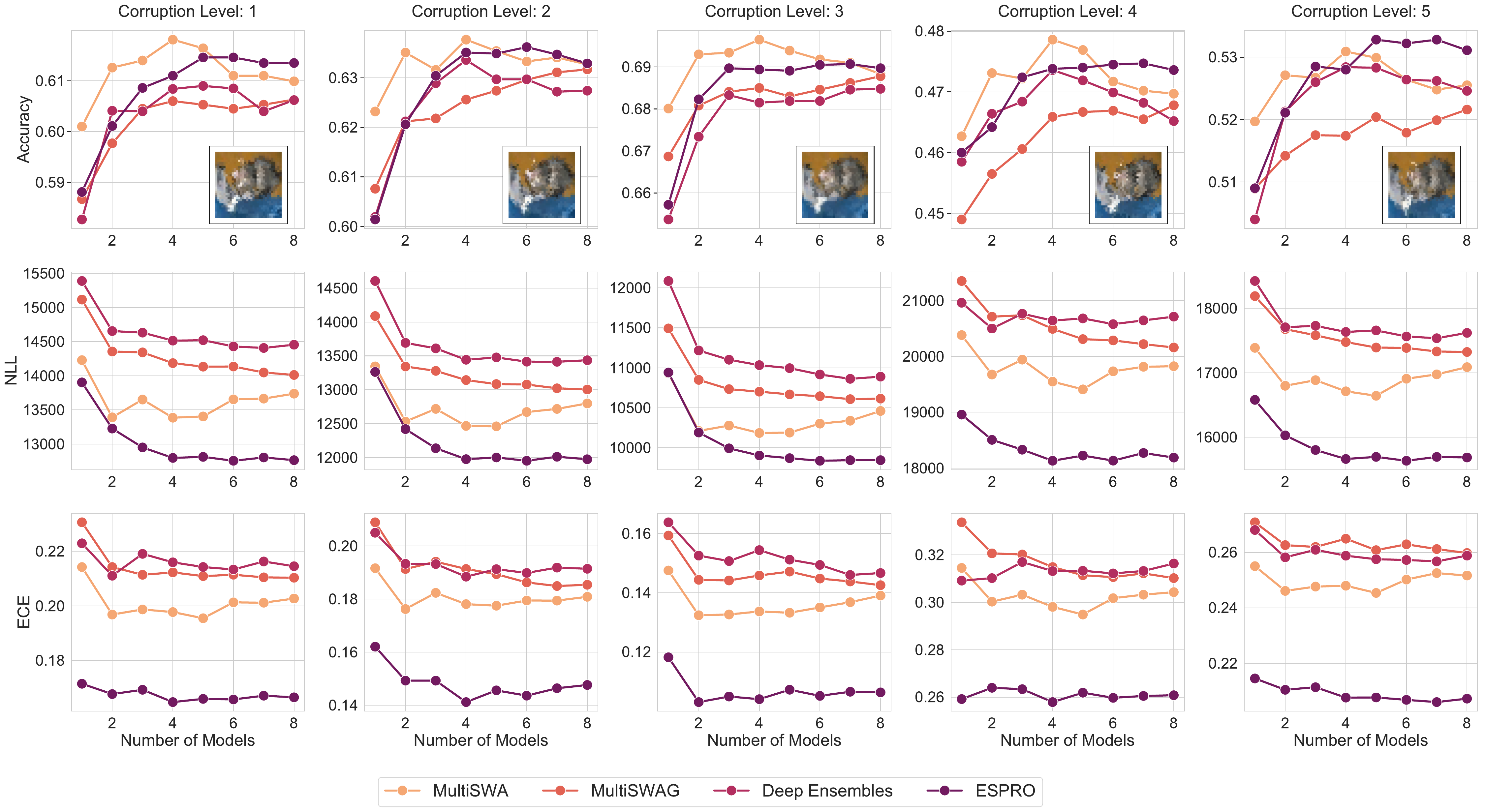}
    \caption{Accuracy, NLL and ECE with increasing intensity of the \textit{glass blur} corruption (from left to right).}
    \label{fig:pixelate_corruption}
\end{figure*}

\begin{figure*}
    \centering
    \includegraphics[width=\linewidth]{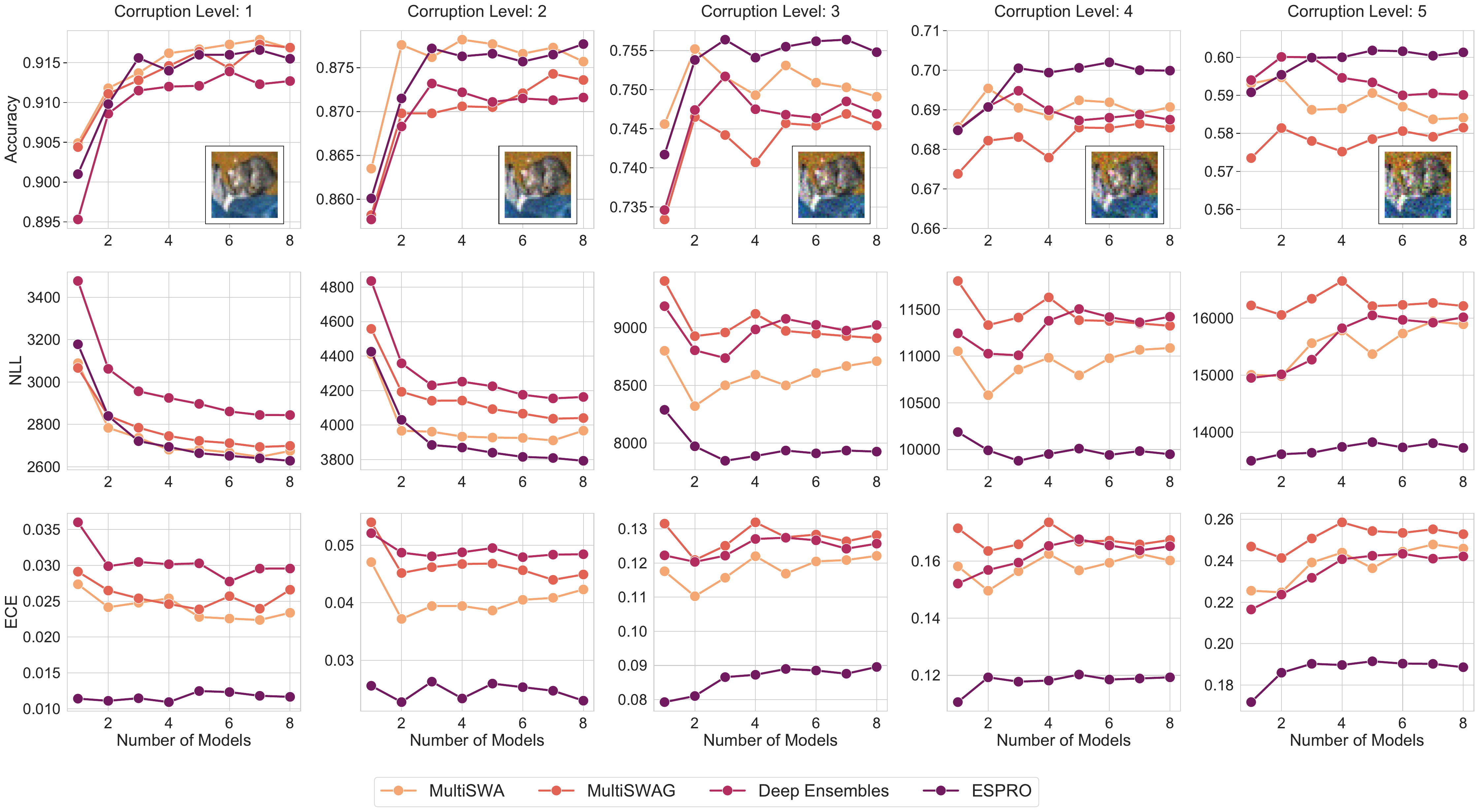}
    \caption{Accuracy, NLL and ECE with increasing intensity of the \textit{shot noise} corruption (from left to right).}
    \label{fig:pixelate_corruption}
\end{figure*}

\begin{figure*}
    \centering
    \includegraphics[width=\linewidth]{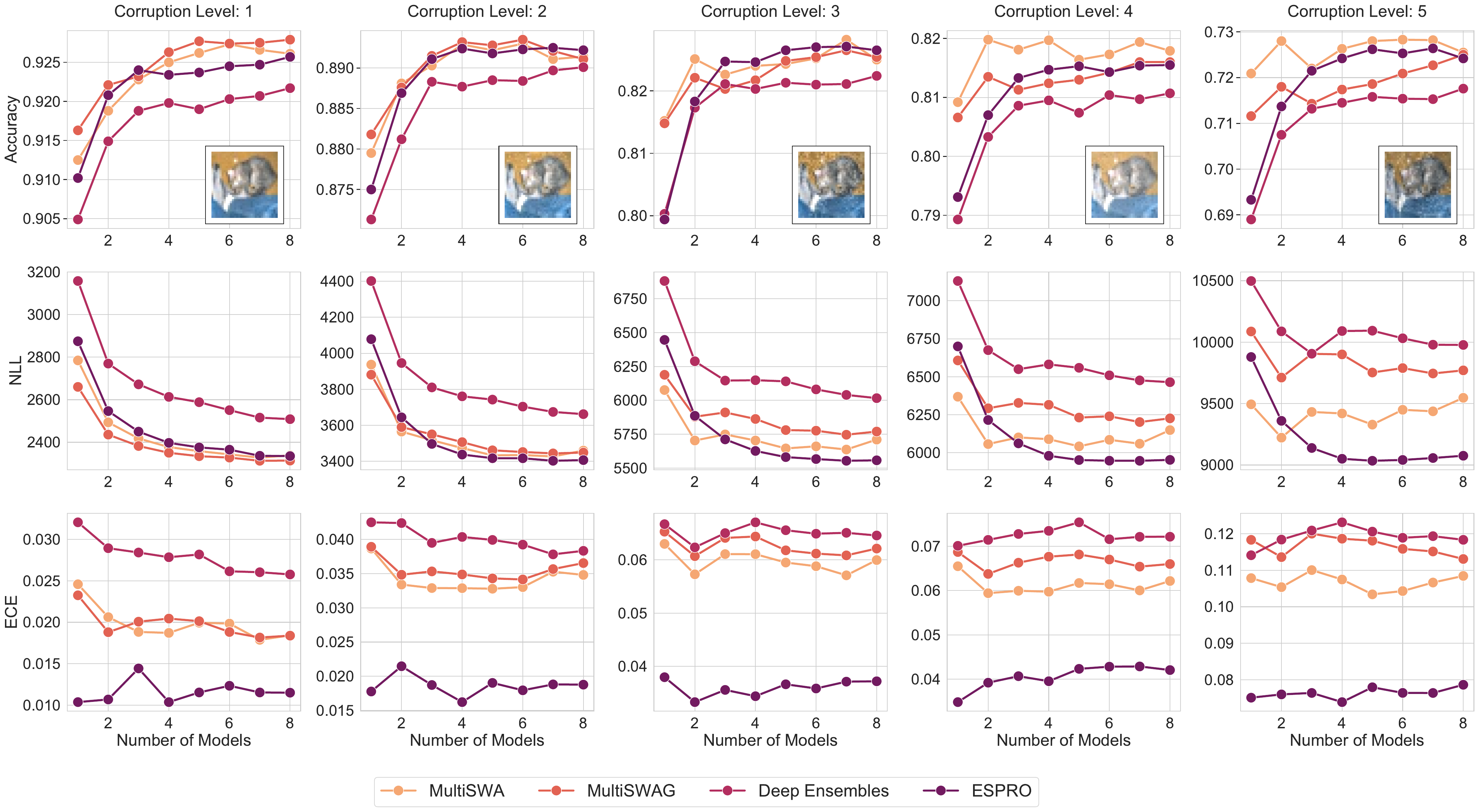}
    \caption{Accuracy, NLL and ECE with increasing intensity of the \textit{frost} corruption (from left to right).}
    \label{fig:pixelate_corruption}
\end{figure*}

\begin{figure*}
    \centering
    \includegraphics[width=\linewidth]{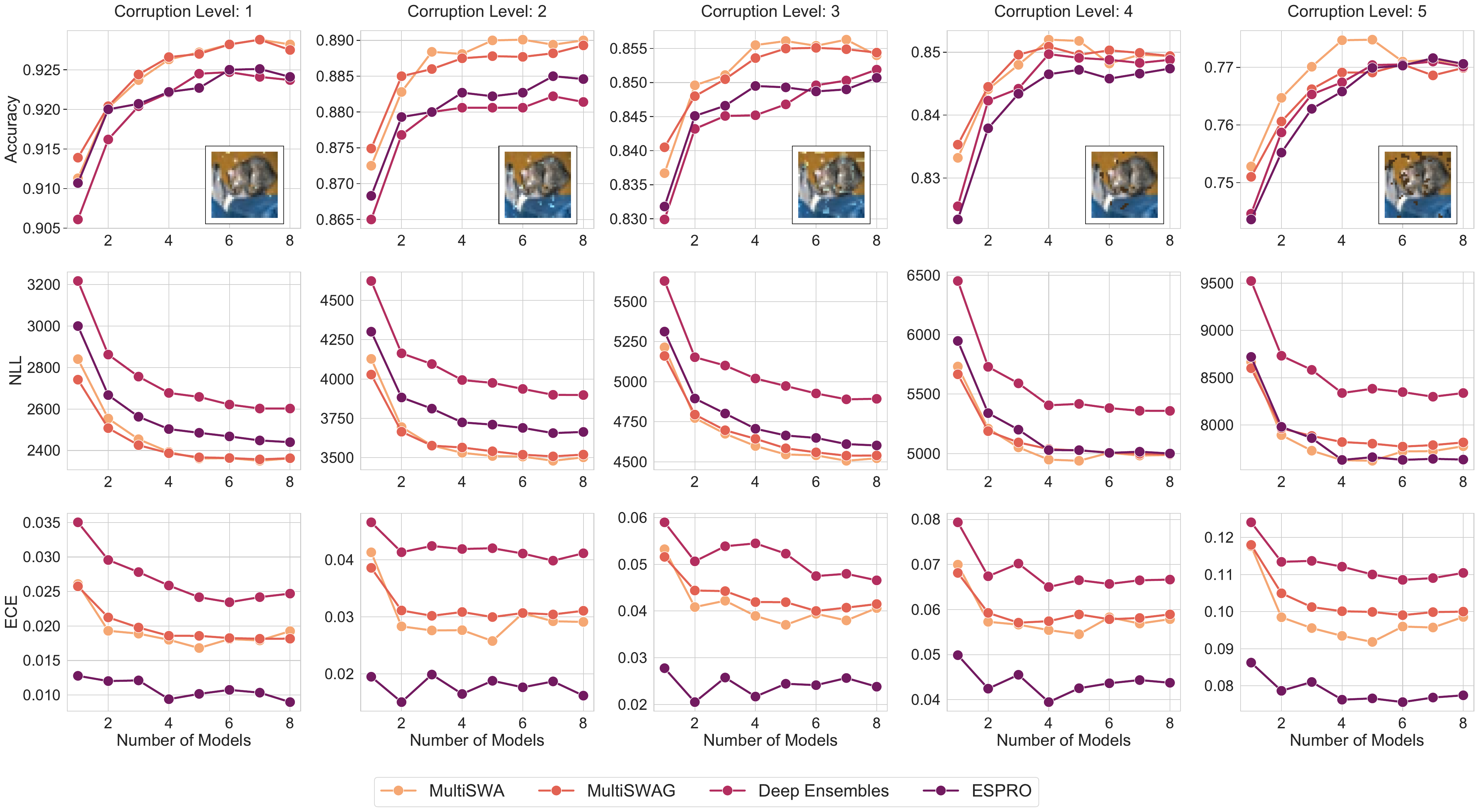}
    \caption{Accuracy, NLL and ECE with increasing intensity of the \textit{spatter} corruption (from left to right).}
    \label{fig:pixelate_corruption}
\end{figure*}

\begin{figure*}
    \centering
    \includegraphics[width=\linewidth]{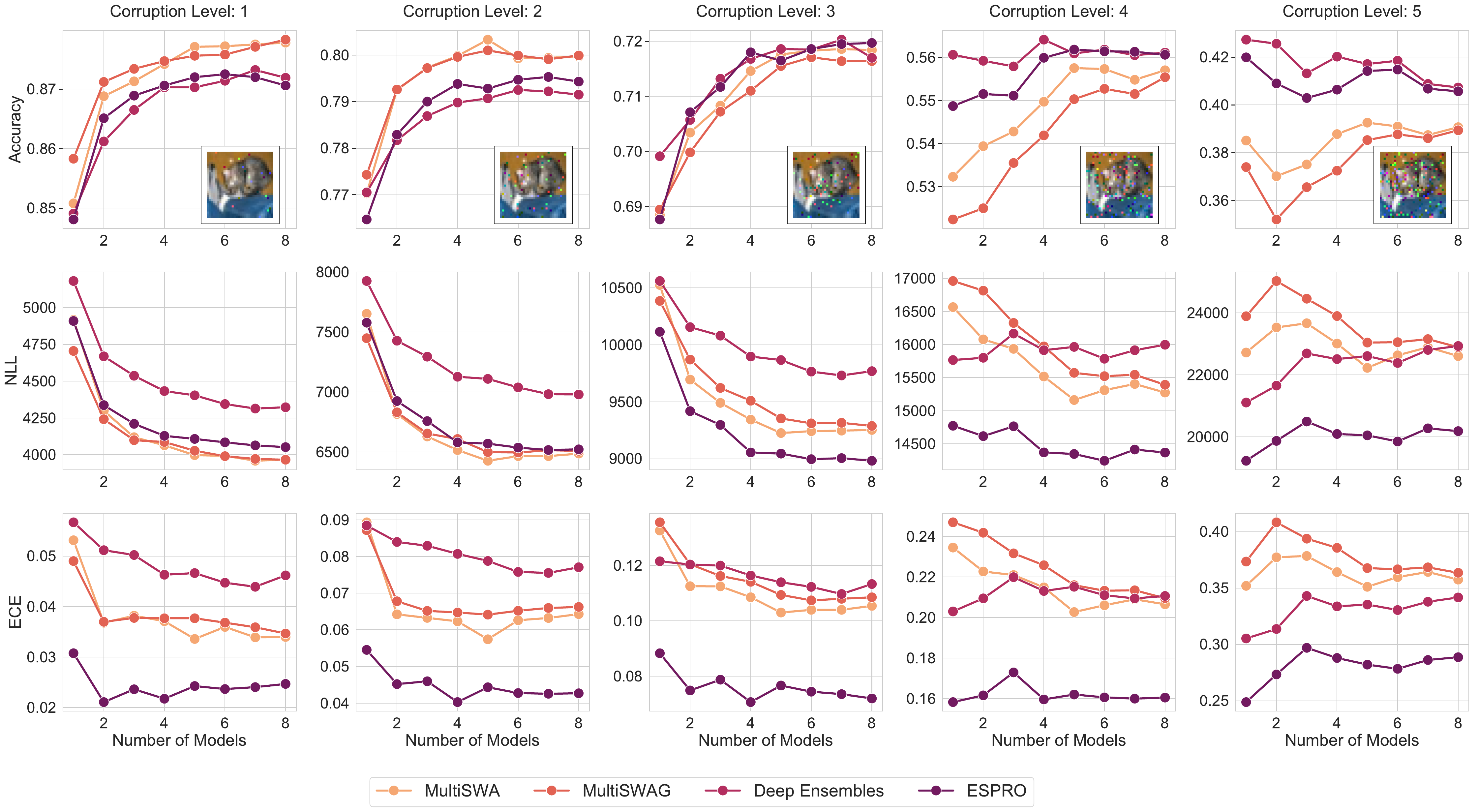}
    \caption{Accuracy, NLL and ECE with increasing intensity of the \textit{impulse noise} corruption (from left to right).}
    \label{fig:pixelate_corruption}
\end{figure*}

\begin{figure*}
    \centering
    \includegraphics[width=\linewidth]{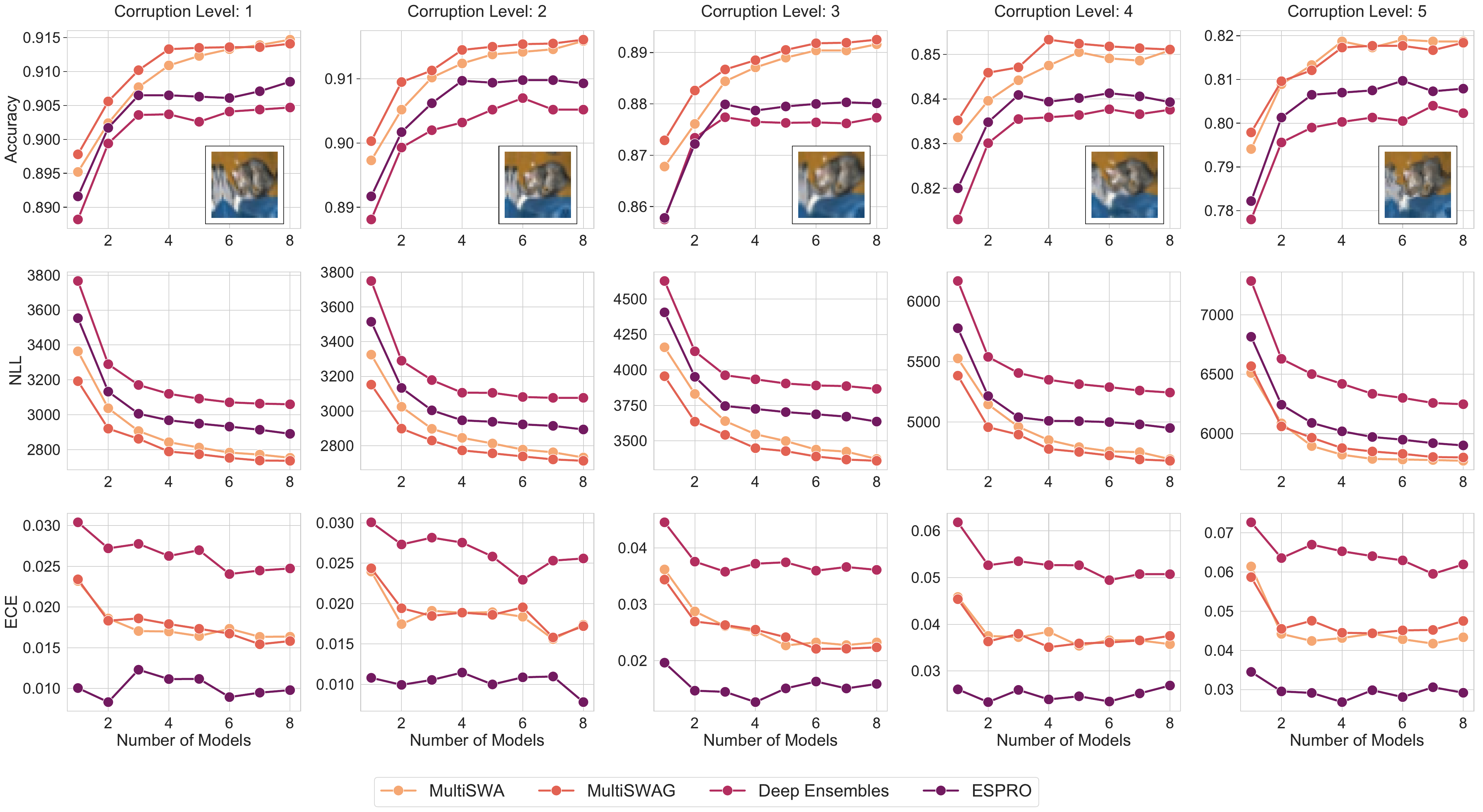}
    \caption{Accuracy, NLL and ECE with increasing intensity of the \textit{elastic transform} corruption (from left to right).}
    \label{fig:pixelate_corruption}
\end{figure*}

\end{document}